%% file: Consistent_model_theory.tex
\documentclass[english]{article}

\usepackage{geometry}
\usepackage[T1]{fontenc}
\usepackage[latin9]{inputenc}
\usepackage{bm}
\usepackage{amsmath,mathtools}
\usepackage{amssymb}
\usepackage[unicode=true,
 bookmarks=false,
 breaklinks=false,pdfborder={0 0 1},colorlinks=false]
 {hyperref}
\hypersetup{colorlinks,citecolor=blue,filecolor=blue,linkcolor=blue,urlcolor=blue}

\makeatletter
\usepackage{amsthm}
\usepackage{cite}  
\usepackage{comment}
\usepackage{natbib}
\usepackage{booktabs}
\usepackage{mathabx}
\usepackage{graphicx}

\usepackage[linesnumbered,ruled,vlined]{algorithm2e}

\SetCommentSty{mycommfont}
\usepackage{algorithmic}

\usepackage{float}
\usepackage{multirow}
\usepackage{dsfont}
\usepackage{tcolorbox}
\usepackage{color}
\definecolor{yxc}{RGB}{255,0,0}
\definecolor{yjc}{RGB}{125,0,0}
\definecolor{ytw}{RGB}{255,69,0}
\definecolor{gen}{RGB}{0,0,200}
\definecolor{zhh}{RGB}{0,100,0}

\allowdisplaybreaks

\newcommand{\defn}{\coloneqq}

\newcommand{\real}{\mathbb{R}}

\newcommand{\mymid}{\,|\,}

\newcommand{\Pdata}{p_{\mathsf{data}}}

\newcommand{\overalpha}{\overline{\alpha}}

\newcommand{\Exs}{\mathbb{E}}

\allowdisplaybreaks

\title{Towards a mathematical theory for \\ consistency training in diffusion models}

\author{Gen Li \footnote{The first two authors contributed equally.} \thanks{Department of Statistics, The Chinese University
 of Hong Kong, Hong Kong.} 
\and  Zhihan Huang \footnotemark[1] \thanks{Department of Statistics and Data Science, Wharton School, University
of Pennsylvania, Philadelphia, PA 19104, USA.} 
\and Yuting Wei\footnotemark[3]  
}

\date{\today}

\makeatother

\begin{document}

\theoremstyle{plain} \newtheorem{lemma}{\textbf{Lemma}}\newtheorem{proposition}{\textbf{Proposition}}\newtheorem{theorem}{\textbf{Theorem}}

\theoremstyle{assumption}\newtheorem{assumption}{\textbf{Assumption}}
\theoremstyle{remark}\newtheorem{remark}{\textbf{Remark}}

\maketitle 

\input{abstract}


\setcounter{tocdepth}{2}
\tableofcontents

%\section{Consistency distillation}
%
%Learn
%\begin{align} \label{eq:distillation}
%\min_f \mathbb{E}_{\substack{z\sim\mathcal{N}(0,I_{d}),x_{0}\sim p_{\mathsf{data}} \\ x_t = \sqrt{\overline{\alpha}_{t}}x_{0}+\sqrt{1-\overline{\alpha}_{t}}z}}
%\Big[\big\|f_t(x_t) - f_{t-1}(\phi_t(x_t))\big\|_{2}^{2}\Big],
%\end{align}
%where
%\begin{align}
%\phi_t(x_t) = \frac{1}{\sqrt{\alpha_t}}\Big(x_t + \frac{1-\alpha_t}{2}s_t(x_t)\Big) =: x_{t-1}.
%\end{align}
%Sample
%\begin{align} \label{eq:sample}
%x_T \sim \mathcal{N}(0, I_d),\quad \text{and}\quad x_1 = f_T(x_T).
%\end{align}
%
%\begin{align}
%f_t(x_t) = \phi_1 \circ \phi_2 \circ \ldots \circ \phi_t(x_t).
%\end{align}

\input{introduction}

\input{preliminary}

\input{main-results}

\input{proof.tex}

\input{discussion.tex}

\section*{Acknowledgements}
Y.~Wei is supported in part by the the NSF grants DMS-2147546/2015447, CAREER award DMS-2143215, CCF-2106778, and the Google Research Scholar Award. 

\appendix

\input{proof-auxiliary.tex}

\input{proof-auxiliary-2}

% \vspace{0.5cm}

\bibliographystyle{apalike}
\bibliography{reference-consistency}

\end{document}

%% file: abstract.tex
\begin{abstract}

Consistency models, which were proposed to mitigate the high computational overhead during the sampling phase of diffusion models, 
 facilitate single-step sampling while attaining the state-of-the-art empirical performance.   
 When integrated into the training phase, consistency models attempt to train a sequence of consistency functions capable of mapping any point at any time step of the diffusion process to its starting point. 
Despite the empirical success, a comprehensive theoretical understanding of consistency training remains elusive. 
This paper takes a first step towards establishing theoretical underpinnings for consistency models. 
	We demonstrate that, in order to generate samples within $\varepsilon$ proximity to the target in distribution (measured by some Wasserstein metric), 
	it suffices for the number of steps in consistency learning to exceed the order of $d^{5/2}/\varepsilon$, 
	with $d$ the data dimension. Our theory offers rigorous insights into the validity and efficacy of consistency models, illuminating their utility in downstream inference tasks.

\end{abstract}

\medskip

%Diffusion models, which generate new data samples by implementing a sequence of steps to revert a Markov process, have demonstrated impressive capabilities in a broad range of generative modeling tasks.  However, they often incur substantially higher computational cost compared to other single-step generative modeling algorithms, thereby limiting their efficiency in real-time applications. 
%, thus allowing for real-time sampling

%% file: introduction.tex
\section{Introduction}

Diffusion models \citep{sohl2015deep,song2019generative,ho2020denoising} have garnered growing interest in recent years due to their impressive capabilities in a wide swath of generative modeling tasks, such as image synthesis, video generation, and audio synthesis \citep{dhariwal2021diffusion,ramesh2022hierarchical,rombach2022high,kong2020diffwave,ho2022imagen,popov2021grad}. 
In comparison with other deep generative models, such as 
%Variational Autoencoder or 
Generative Adversarial Networks, which oftentimes suffer from training instability and mode collapse, diffusion models are capable of generating high-fidelity samples based on learning the gradient of the log-density function or the score function. 
On a high level, diffusion models concentrate on two processes: a forward Markov process that gradually degrades data into noise, and a reverse-time stochastic or deterministic process that starts from pure noise, performs iterative denoising to generate new data that resemble true data samples in distribution. 
Interestingly, while the forward process is straightforwardly often designed by progressively injecting more noise into the data samples, 
it is feasible to revert the process and ensure (almost) matching marginals as the forward process, 
as long as faithful score function estimates are obtainable \citep{anderson1982reverse,haussmann1986time}.

Nevertheless, 
given that diffusion models generate new data by implementing a sequence of steps in the reverse process (with each step computing the score function by evaluating a large neural network),  
they often incur substantially higher computational cost compared to other single-step generative modeling algorithms, 
thereby limiting their sampling efficiency in real-time applications.  
To remedy this issue, there has been an explosion of efforts in developing acceleration procedures to speed up the sampling process in diffusion generative modeling (e.g.~\cite{song2020improved,lu2022dpm,lu2022dpmv2,zhao2023unipc,zhang2022fast,xue2023sa,luhman2021knowledge,salimans2022progressive,song2023consistency,li2024accelerating}).
Among these efforts, training-based methods, 
exemplified by progressive distillation and consistency models hold great promises in 
producing samplers that are computationally efficient and ready for real-time implementation without sacrificing sampling fidelity \citep{salimans2022progressive,meng2023distillation,sun2023accelerating,song2023consistency}. 

In this paper, our focal point  
is the consistency model, 
 which was originally proposed by \cite{song2023consistency} and claims the state-of-the-art performance.  
 %Built upon a reverse process constructed through the probability flow ordinary differential equations (ODEs), 
 In a nutshell, 
 the consistency model seeks to learn a function that is able to map any point at any time step of the diffusion process to the process' starting point (the end corresponding to the data distribution).  
 In the sampling phase, the consistency model enables sample generation with only a single evaluation of the neural network. 
The surprising efficacy of consistency models has been demonstrated in various image datasets, including CIFAR-10, ImageNet $64\times 64$, LSUN $256\times 256$, and also video generation 
\citep{song2023consistency,wang2023videolcm}, to name just a few. 
This approach has received considerable recent attention, 
covering various extensions  
(e.g.~\cite{song2023improved,kim2023consistency}) as well as applications beyond generative models (e.g.~reinforcement learning \citep{ding2023consistency}).

Despite the aforementioned mind-blowing empirical successes, 
however, 
a theoretical understanding of consistency models remains elusive even in the most basic setting. 
In light of the flexibility and versatility of the consistency model idea (which only requires enforcing some self-consistency conditions), 
%consistency models allow great flexibilities in the training process as long as the self-consistency condition satisfies, 
establishing theoretical underpinnings for these models not only provides rigorous justifications for their validity, 
but also yields practical implications in downstream inference tasks by providing theoretical benchmarks to compare different training strategies.  
However, the challenge in establishing theoretical performance guarantees lies in understanding the role of consistency enforcement in preserving the sampling fidelity.

% interplay between the sampling fidelity and consistency along the trajectory. 

\paragraph{An overview of our contributions.}
In this paper, 
we take a first step towards establishing theoretical support for consistency models, 
focusing on consistency training (namely, applying the consistency model idea from the training stage). 
More specifically, we consider a consistency training paradigm that recursively learns a sequence of functions $\{f_t\}_{1\leq t \leq T}$, 
in the hope that the ultimate sampling process can be readily completed by evaluating $f_{T}(X_T)$ with $X_{T} \sim \mathcal{N}(0,I_d).$
Our theory reveals that: it is sufficient for consistency training to take a number of steps exceeding the order of
	\begin{align}
		\frac{d^{5/2}}{\varepsilon}
	\end{align}
	up to some logarithmic factor 
	in order to generate samples that are $2\varepsilon$ close in distribution to the target data distribution (measured by the Wasserstein metric). 
	Here, $d$ denotes the dimension of the target distribution, and we omit the logarithm factors and dependence on other universal constants.
	% This result offers an explicit characterization on the dependence of the Lipschitz constant as well as the dimension of the problem. 
In other words, it tells us how many steps need to be included in the training stage in order to enable one-shot sampling that achieves the desirable sampling fidelity.

%We highlight several of our main contributions below. 

%\begin{itemize}
%	\item Our theory guarantees that, it is sufficient for consistency training to take a number of steps exceeding
%	\begin{align}
%		\widetilde{O} \Big(\frac{d^{5/2}}{\varepsilon}\Big),
%	\end{align}
%	to generate samples that are $2\varepsilon$ close to the target distribution in Wasserstein metric. 
%	Here, $d$ denotes the dimension of the target distribution, and we omit the logarithm factors and dependence on other universal constants.
%	% This result offers an explicit characterization on the dependence of the Lipschitz constant as well as the dimension of the problem. 
%
%	\item 
%\end{itemize}

% This paper takes a first step towards building the theoretical foundation of consistency models with an emphasis on the non-asymptotic performances of consistency training with explicit dependence on the salient parameters. 

% This paper develops the first mathematical framework to analyze consistency training in diffusion models. 

\paragraph{Notation.}
We introduce a couple of notation to be used throughout this paper.  
Given two probability measures $\mu$ and $\nu$ on $\real^{d}$, we denote by $\mathcal{C}(\mu,\nu)$ the set of all couplings of $\mu$ and $\nu$ (i.e., all joint distributions $\gamma(x,y)$ whose marginals coincide with $\mu$ and $\nu$, respectively). 
The Wasserstein distance of order $q$ between these two distributions is defined as 
\begin{align}
\label{eqn:wasserstein-p}
	W_q(\mu, \nu) \defn \bigg(\inf_{\gamma \in \mathcal{C}(\mu,\nu)} \mathop{\mathbb{E}}\limits_{(x,y)\sim \gamma}\big[ \|x - y\|_2^q \big] \bigg)^{1/q}, 
\end{align}
and we often employ $W_q(X, Y)$ for random variables $X$ and $Y$ to denote the Wasserstein distance between distributions of $X$ and $Y.$ 
In addition, given any two functions $f(d,T)$ and $g(d,T)$, we write $f(d,T)\lesssim g(d,T)$ or $f(d,T)=O( g(d,T) )$ (resp.~$f(d,T)\gtrsim g(d,T)$) if there exists some universal constant $C_1>0$ such that $f(d,T)\leq C_1 g(d,T)$ (resp.~$f(d,T)\geq C_1 g(d,T)$) for all $d$ and $T$. 
Furthermore, the notation $\widetilde{O}(\cdot)$ is defined analogously to $O(\cdot)$ except that the logarithmic dependency is hidden. 
Given a matrix $M \in \real^{d\times d}$, we denote $\|M\|$ as the operator norm of $M.$

% \vspace{2cm}

%% file: preliminary.tex
\section{Preliminaries}

In this section, we introduce the basics of diffusion generative modeling and consistency models. 
While the consistency model was originally motivated to accelerate the probability flow ODE sampler and distill information from a pre-trained model, 
the idea of promoting consistency along the trajectory can be incorporated directly into the training stage, which we focus on in this paper. 

%training a consistency model does not necessarily rely on a pre-trained diffusion model, making consistency models an independent class of generative models. 
%

\subsection{Diffusion-based generative models}

% \yuting{start directly from probability flow ODE?}

\paragraph{Forward process.} 
As briefly mentioned above, in diffusion generative models, one starts from a forward process and progressively perturbs the data into pure noise, 
where the noise distribution is often chosen to be Gaussian. 
The forward process is often modeled as solution to an It\^{o} stochastic differential equation (SDE)
\begin{align}
	\mathrm{d}X_{t} = f(X_t, t) \mathrm{d}t + g(t)\mathrm{d}W_t \qquad (0\leq t\leq T),
\end{align}
where $W_{t}$ corresponds to a standard Brownian motion, $f(\cdot, t): \real^{d} \to \real^{d}$ is a vector-valued function that determines the drift of this process, and $g(\cdot): \real \to \real$ is a function that adjusts the variance of the injected noise. 
We shall adopt the notation $q_{t} := \textsf{Law}(X_t)$ throughout to represent the distribution of $X_{t}$ in this forward process. 
In particular, $q_{0}:= \textsf{Law}(X_0)$ is our target distribution to generate samples from, and it is also frequently denoted by $\Pdata$. 
A popular special case that motivates DDPM and DDIM algorithms \citep{song2020denoising,ho2020denoising,nichol2021improved} is to take $f(X_t, t) = - \frac{1}{2} \beta(t) X_t$ and $g(t) = \sqrt{\beta(t)}$ for some function $\beta(\cdot)$ (which can be interpreted as determining the learning rate schedule). The SDE defined above then reduces to 
\begin{align}
	\mathrm{d}X_{t}= - \frac{1}{2} \beta(t) X_t \mathrm{d}t+ \sqrt{\beta(t)}\,\mathrm{d}W_{t} \quad (0\leq t\leq T),
	\qquad X_{0}\sim\Pdata.
	\label{eq:forward-SDE}
\end{align}

Given the continuous-time nature of the above forward process, 
it would oftentimes be helpful to look at the discrete-time counterpart instead. 
More specifically, consider the following discrete-time random process:    
\begin{subequations}
\label{eq:forward-process}
\begin{align}
	X_0 &\sim \Pdata,\label{eq:forward-process-1}\\
	X_t &= \sqrt{1-\beta_t}X_{t-1} + \sqrt{\beta_t}\,W_{t}, \qquad 1\leq t\leq T,\label{eq:forward-process-2}
\end{align}
\end{subequations}
with $T$ representing the total number of steps. 
Here, we denote by $\{\beta_t \} \subseteq (0,1)$  the prescribed learning rates that control the strength of the noise injected at each step, and $\{W_t\}_{1\leq t\leq T}$  a sequence of independent noise vectors drawn from $W_{t} \overset{\mathrm{i.i.d.}}{\sim} \mathcal{N}(0, I_d)$. 
If we further denote 
\begin{align}
\label{eqn:alpha-t}
	\alpha_t\coloneqq 1 - \beta_t, 
	\qquad \overline{\alpha}_t \coloneqq \prod_{k = 1}^t \alpha_k ,\qquad 1\leq t\leq T,
\end{align}
one can write 
\begin{align}
\label{eqn:Xt-X0}
	X_t = \sqrt{\overalpha_t} X_{0} + \sqrt{1-\overalpha_t} \,\overline{W}_{t}
	\qquad \text{for some } \overline{W}_{t}\sim \mathcal{N}(0,I_d).
\end{align}
In practice, $\overalpha_{T}$ is oftentimes chosen to be vanishingly small (as long as $T$ is large enough), so as to make sure that the distribution $q_T$ of $X_T$ is approximately $\mathcal{N}(0,I_d)$.

\paragraph{Reverse process.} 

Reversing the above process in time leads to a process that transforms noise into samples with distribution approximately equal to $\Pdata$, which is how diffusion models  generate data.
In particular, by the classical results \citep{anderson1982reverse,haussmann1986time}, a reverse-time SDE corresponding to \eqref{eq:forward-SDE} obeys 
\begin{align}
\mathrm{d}Y_{t}^{\mathsf{sde}}  
& 
= \Big(\frac{1}{2} \beta(T-t) Y_{t}^{\mathsf{sde}}  + \beta(T-t)\nabla\log q_{T-t} (Y_{t}^{\mathsf{sde}} )\Big)\mathrm{d}t + \sqrt{\beta(T-t)}\mathrm{d}Z_{t}\quad(0\leq t\leq T) 	
\label{eq:reverse-SDE}
\end{align}
with $Y_{0} \sim q_{T}$ and $Z_t$ being a standard Brownian motion, which satisfies $Y_{T-t}^{\mathsf{sde}} \overset{\mathrm{d}}{=} X_t$. Let $p_{t}:= \textsf{Law}(Y_t)$ be the distribution of $Y_{t}$.
Evidently, to implement such a process, it requires obtaining faithful estimates of the score function \footnote{For notational convenience, we also adopt the shorthand notation $\nabla \log q_{t}(X)$ to denote the score function (by suppressing the dependency on $X$).} 
\begin{align}
	s_t(X) := \nabla_{X} \log q_{t}(X).
\end{align}

The existence of this reverse-time SDE has formed the underlying rationale for the design of the popular sampler called the Denoising Diffusion Probabilistic Model (DDPM).

Another popular sampler, called the Denoising Diffusion Implicit Model (DDIM) \citep{karraselucidating,song2020denoising,song2020score}, leverages upon 
the so-called probability flow ODE. More precisely, consider the following ODE 
\begin{align}
	\mathrm{d}Y_{t}^{\mathsf{ode}} 
	& 
	= 
	\frac{1}{2} \beta(T-t)\Big(Y_{t}^{\mathsf{ode}} + \nabla\log q_{T-t} (Y_{t}^{\mathsf{ode}})\Big)\mathrm{d}t
	\quad(0\leq t\leq T),\qquad Y_{0}^{\mathsf{ode}}\sim q_{T},
	\label{eq:prob-flow-ODE}
\end{align}
which again yields matching marginal distributions for $X_{t}$ as 
\begin{align*}
	Y_{T-t}^{\mathsf{ode}}  \, \overset{\mathrm{d}}{=} \, X_t, \qquad 0\leq t\leq T. 
\end{align*}
Generating a data sample from this ODE again only requires reliable estimation of the score function. 
It is noteworthy that this deterministic ODE-based approach is often faster than the SDE-based approach \citep{song2020score}, 
which has also been justified in theory \citep{li2023towards}. 

%We remark that since this process is determined by an ODE, which is deterministic in nature, this sampling approach is oftentimes less time-consuming compared to the SDE based approaches. 

We note that the probability flow ODE considered here in \eqref{eq:prob-flow-ODE} is slightly different from the one in \citet{song2023consistency}, the latter of which takes the form
\begin{align}
	\mathrm{d}Y_{t} = -t\nabla\log q_{t} (Y_{t}) \mathrm{d}t 
\end{align}
and corresponds to the forward process $\mathrm{d}X_{t} = \sqrt{2t} \mathrm{d}W_t.$
In particular, if the covariance of $X_{0}$ is equal to $I_d$, then $q_{T}$ is close to a Gaussian distribution $\mathcal{N}(0, T^2I_d)$ (so that the covariance explodes),
whereas in the process~\eqref{eqn:Xt-X0}, the covariance for $X_{t}$ is preserved and equals $I_d$ throughout the trajectory.

Finally, we note that 
recent years have witnessed remarkable theoretical advances towards understanding the sampling performance of diffusion models. 
A highly incomplete list includes \cite{block2020generative,de2021diffusion,liu2022let,de2022convergence,lee2023convergence,pidstrigach2022score,chen2022sampling,tang2023diffusion,benton2023linear,chen2022improved,chen2023restoration,tang2024contractive,li2023towards}. In particular, the recent works \cite{chen2022sampling,chen2022improved,benton2023linear,chen2023restoration,chen2023probability,li2023towards} 
have established the convergence rates of both the DDPM and DDIM samplers, as well as their stability against $\ell_2$ score estimation errors.

\subsection{Consistency training}

While the probability flow ODE approach already achieves much faster sampling compared to the DDPM sampler, 
it still requires a large number of steps (or equivalently, a large number of neural network evaluations) and does not yet meet the demand for real-time sample generation. 
This motivates the development of the consistency model  as a means to accomplish sampling in one step \citep{song2023consistency}.

Specifically, given a solution trajectory $\{x_{t}\}_{t\in [\epsilon, T]}$ of the probability ODE in \eqref{eq:prob-flow-ODE}, 
a consistency function is a parametrized function (parameterized by $\theta$) designed to achieve
\begin{align}
	f_\theta : (x_t, t ) \overset{\text{ideally}}{\longrightarrow} x_{\epsilon} \qquad \text{ for all } t\in [\epsilon, T],
\end{align}
which maps a point $x_{t}$ at time $t$ back to the desired sample $x_{\epsilon}$.
Therefore, given a well-trained consistency model $f_\theta$, in the sampling phase, instead of recursively applying denoising function $p_\theta(x_{t-1} \mid x_t)$ as the reverse diffusion process in diffusion model, 
it suffices to evaluate $f_\theta (\widehat{x}_T, T)$ once to produce an approximation of $\widehat{x}_{\epsilon}.$
By doing so, one forward pass through the consistency model (or one evaluation of the neural network) suffices to generate a sample that mimics the target distribution.

When the consistency approach is integrated into the training phase, 
it entails an iterative procedure to find suitable parameterization $\theta.$ 
%Without a pre-trained diffusion model, 
More specifically, the idea put forward by \cite{song2023consistency} is to minimize a certain consistency training objective over the parameter $\theta$ where 
\begin{align}
	\mathop{\text{minimize}}\limits_{\theta} \quad \mathcal{L}_{\theta^-}(\theta) 
	:= \mathbb{E}\Big[\lambda(t_n) \cdot \textsf{dist}\big(f_\theta(X_0 + t_{n+1}Z, t_{n+1}), 
	f_{\theta^-}(X_0 + t_{n}Z,t_n)\big)\Big],
\end{align}
where the time horizon $[\epsilon, T]$ is discretized into $N- 1$ sub-intervals, with boundaries $t_{1} = \epsilon < t_{2} < \ldots < t_{N} = T$.\footnote{The exact formulas for $\{t_{i}\}_{1\leq i\leq N}$ can be found in \cite{song2023consistency}.}
Here, $\textsf{dist}(\cdot, \cdot)$ is some distance measure between two vectors in $d$ dimension, $\lambda(\cdot)$ is some weighting function, and $\theta^{-}$ is some moving average of $\theta$ during the course of training. 
The expectation is taken over $X_0 \sim \Pdata$, $Z\sim \mathcal{N}(0, I_d)$ and $n$ drawn uniformly from $\{0,1,\ldots, N\}.$
% \yuting{is this correct? should we directly state what we are analyzing? TBD}

\paragraph{This paper: iterative consistency training.} Motivated by the above strategy, the current work proposes to iteratively learn a sequence of consistency functions $\{f_t\}_{1\leq t\leq T}$ with $f_{t}: \real^{d} \to \real^{d}$, without making explicit the parameterization $\theta$. Here, each $f_{t}$ is allowed to be selected from a function class $\mathcal{F}$ (e.g., it can be specified by some large neural network).  
Taking $f_1(x) \equiv x$, we propose to learn $f_t$ for each $2 \le t \le T$ by minimizing the following objective:
\begin{align} 
\label{eq:CT}
	f_t \approx 
	\mathop{\arg \min}\limits_{f \in \mathcal{F}} ~\mathbb{E}\Big[\big\|f(\sqrt{\overline{\alpha}_{t}}X_{0}+\sqrt{1-\overline{\alpha}_{t}}Z) - f_{t-1}(\sqrt{\overline{\alpha}_{t-1}}X_{0}+\sqrt{1-\overline{\alpha}_{t-1}}Z)\big\|_{2}^{2}\Big],
\end{align}
where the expectation is taken with respect to $Z\sim\mathcal{N}(0,I_{d})$ and $X_{0}\sim p_{\mathsf{data}}$. 
By computing $f_t$ as in \eqref{eq:CT} step by step, we ensure that $f_t(X_t)\approx f_{t-1}(X_{t-1}) \approx \dots \approx X_1$. 
In the sampling stage, to generate a new sample from $\Pdata$, it is sufficient to draw a point $X_{T} \sim \mathcal{N}(0, I_d)$ and then output $f_{T}(X_T)$.

%% file: main-results.tex
\section{A non-asymptotic convergence theory for consistency training}

\subsection{Assumptions and setup}

Before delving into our main results, let us introduce several notation and terminologies. 
For any $0 < \overline{\alpha} < 1$, we denote
\begin{align}
	X(\overline{\alpha}) = \sqrt{\overline{\alpha}}X_{0}+\sqrt{1-\overline{\alpha}}Z,
	\quad \text{ and } \quad
	s_{\overline{\alpha}}(x) = \nabla \log p_{X(\overline{\alpha})}(x),
\end{align}
where $X_0 \sim \Pdata$ and $Z \sim \mathcal{N}(0, I_d).$
Throughout this paper, we shall abuse the notation and let $$s_{t}(x) := s_{\overline{\alpha}_t}(x),$$ 
with $\overline{\alpha}_t$ defined in expression~\eqref{eqn:alpha-t}.

With these definitions in mind, if we define 
\begin{align}
 	\Phi_{t \to k}(x) \defn g_t(x,\overline{\alpha}_k)
\end{align}
with
\begin{align}
\label{eq:ODE}
\frac{\partial g_t(x,\overline{\alpha})}{\partial \overline{\alpha}} 
= \frac{1}{2\overline{\alpha}} \Big(g_t(x,\overline{\alpha}) + s_{\overline{\alpha}}(g_t(x,\overline{\alpha}))\Big),
\qquad\text{and }g_t(x,\overline{\alpha}_t) = x,
\end{align}
then we claim that the  probability flow ODE in \eqref{eq:prob-flow-ODE} satisfies $\Phi_{t \to k}(X_t) \overset{\mathsf{d}}{=} X_k$.
In order to prove this relation, it suffices to make use of the ODE given in \eqref{eq:prob-flow-ODE} and the observation that 
$\mathrm{d}\overline{\alpha}_{t} = - \overline{\alpha}_{t} \beta_t \mathrm{d}t.$ 
In fact, property~\eqref{eq:ODE} holds true for the entire ODE trajectory in the sense that $g_t(X_t,\overline{\alpha}) \overset{\mathsf{d}}{=} X(\overline{\alpha})$.
%In addition, define
%\begin{align}
%\Phi_{t \to k}(x) = \phi_{k+1} \circ \ldots \circ \phi_{t-1} \circ \phi_t(x),
%\end{align}
%where
%\begin{align}
%\phi_t(x) = \frac{1}{\sqrt{\alpha_t}}\Big(x + \frac{1-\alpha_t}{2}s_t(x)\Big).
%\end{align}
%Note that $\Phi_{t \to k}$ is the map by discretizing $\Phi_{t \to k}^{\star}$.
For notational simplicity, we shall denote
\begin{align}
	\phi_{t} := \Phi_{t \to t-1} \quad \text{ and } \quad \Phi_{t} := \Phi_{t \to 1},
\end{align}
and it satisfies
\begin{align*}
	\Phi_{t \to k}(x) = \Phi_{k+1 \to k}\circ\cdots \Phi_{t-1 \to t-2}\circ\Phi_{t \to t-1}(x)
	 = \phi_{k+1}(\phi_{k+2}(\cdots \phi_t(x)\cdots)).
\end{align*}

The consistency functions $\{f_t\}_{1\leq t \leq T}$  are trained according to \eqref{eq:CT} without a pre-trained backward process. To measure the quality of the output samples which are distributed according to $f_{T}(X_T)$, we consider the Wasserstein distance $\mathcal{W}_1\big(f_T(X_T), X_1)$ with $X_T\sim \mathcal{N}(0,I_d)$. 
Our main result is established under the following two assumptions.

\begin{assumption} 
\label{ass:Lipschitz}
Assume that for $1 \le k < t \le T$, $\Phi_{t \to k}$ is $L_f$-Lipschitz continuous such that
\begin{align}
\big\|\Phi_{t \to k}(x) - \Phi_{t \to k}(y)\big\|_2 \le L_f\|x - y\|_2.
\end{align}
%Moreover, suppose that for $1 \le t \le T$, $s_{t}$ is $L_s$-Lipschitz continuous such that
%\begin{align}
%\big\|s_{t}(x) - s_{t}(y)\big\|_2 \le L_s\|x - y\|_2.
%\end{align}
\end{assumption}

\begin{assumption} 
\label{ass:estimation}
Given a function class $\mathcal{F}$,  suppose there exist $\varepsilon, \varepsilon_{\mathcal{F}} > 0$ such that the functions $\{f_t\}_{1\leq t \leq T}$ trained according to \eqref{eq:CT} satisfy
	\begin{subequations}
\begin{align}
&\sum_{t=1}^T\mathbb{E}\Big[\big\|f_t(\sqrt{\overline{\alpha}_{t}}X_{0}+\sqrt{1-\overline{\alpha}_{t}}Z) - f_t^{\mathcal{F}}(\sqrt{\overline{\alpha}_{t}}X_{0}+\sqrt{1-\overline{\alpha}_{t}}Z)\big\|_{2}\Big] \le \varepsilon,\\
&\sum_{t=1}^T\mathbb{E}\Big[\big\|f^{\mathcal{F}}_t(\sqrt{\overline{\alpha}_{t}}X_{0}+\sqrt{1-\overline{\alpha}_{t}}Z) - f_t^{\star}(\sqrt{\overline{\alpha}_{t}}X_{0}+\sqrt{1-\overline{\alpha}_{t}}Z)\big\|_{2}\Big] \le \varepsilon_{\mathcal{F}},
\end{align}
	\end{subequations}
where
\begin{subequations}
\begin{align}\label{eq:true-dis}
f_t^{\mathcal{F}} \defn& \arg\min_{f \in \mathcal{F}} \mathbb{E}
\Big[\big\|f(\sqrt{\overline{\alpha}_{t}}X_{0}+\sqrt{1-\overline{\alpha}_{t}}Z) - f_{t-1}(\sqrt{\overline{\alpha}_{t-1}}X_{0}+\sqrt{1-\overline{\alpha}_{t-1}}Z)\big\|_{2}^{2}\Big],\\
f_t^{\star} \defn&\mathbb{E}\Big[f_{t-1}\big(\sqrt{\overline{\alpha}_{t-1}}X_{0}+\sqrt{1-\overline{\alpha}_{t-1}}Z\big)\big|\sqrt{\overline{\alpha}_{t}}X_{0}+\sqrt{1-\overline{\alpha}_{t}}Z\Big].
\end{align}
\end{subequations}
\end{assumption}

In words, Assumption~\ref{ass:Lipschitz} requires the mappings from $X_{t}$ to $X_{k}$ to be Lipschitz continuous for every $k$ and $t$. 
Assumption~\ref{ass:estimation} is concerned with two sources of errors in the training process: {(i)} $\varepsilon$ controls the estimation error of the consistency functions $\{f_{t}\}_{1\leq t \leq T}$ we have obtained; 
{(ii)} $\varepsilon_{\mathcal{F}}$ corresponds to the approximation error of restricting the consistency functions to lie within some fixed function class $\mathcal{F}$, where $\varepsilon_{\mathcal{F}} = 0$ for function classes with large capacity (or representation power) like neural networks.
We remark that the optimization step \eqref{eq:CT}  for obtaining~$\{f_{t}\}_{1\leq t \leq T}$ is typically accomplished through proper training of large neural networks.
Given the complexity of developing an end-to-end theory, we adopt the common divide-and-conquer strategy and decouple the training phase with the sampling phase. In the sequel, we shall focus our attention on quantifying the sampling fidelity, assuming small estimation/optimization errors in the training phase. 

\paragraph{Target data distribution.}
To streamline our main proof, we impose an additional constraint on the target data distribution, namely, 
\begin{align}
\label{eqn:boundness}
	\mathbb{P}(\|X_0\|_2 \le T^{c_R}) = 1,
\end{align}
where $X_0 \sim \Pdata$ and $c_R>0$ is some arbitrarily large constant.
This assumption covers a broad family of data distribution with polynomially large support size. 
We remark that this constraint can be replaced by some careful assumptions on the tail probability of the target data distribution, and the resulting proof is expected to be similar.

\paragraph{Learning rate schedule.}
Finally, let us specify the learning rate schedule $\{\alpha_t\}_{1\leq t \leq T}$ we would like to employ during  consistency training~\eqref{eq:CT}. 
For some large enough numerical constants $c_0,c_1 > 0$, we set
\begin{subequations}
	\label{eqn:alpha-t}
	\begin{align}
	\beta_{1} & =1-\alpha_{1} = \frac{1}{T^{c_0}};\\
	\beta_{t} & =1-\alpha_{t} = \frac{c_{1}\log T}{T}\min\bigg\{\beta_{1}\Big(1+\frac{c_{1}\log T}{T}\Big)^{t},\,1\bigg\}, \qquad~~ 2\leq t\leq T\\
	\overline{\alpha}_t &= \prod_{i=1}^{t} \alpha_i, \qquad~~ 2\leq t\leq T.
	\end{align}
\end{subequations}
Note that such scheduled learning rates have been employed in the prior work \citet{li2023towards} to achieve the desired convergence guarantees.  
% enssure that almost pure Gaussian noise at last. 
%the noise we add in each step is sufficiently small compared to  noise level, and we 
% Accordingly, it is fair for us to consider denoising process begin with $X_T$ in place of $\mathcal{N}(0,I_d)$.
A couple of other useful properties about these learning rates are provided in Section~\ref{sec:main-proof}. 
% , \eqref{eqn:properties-alpha-proof}. 

\subsection{Main results}
We are now positioned to state our main theoretical guarantees for consistency training.
\begin{theorem}
\label{thm:main}
Suppose the learning rates are selected according to \eqref{eqn:alpha-t} and the target distribution satisfies property~\eqref{eqn:boundness}.
Under Assumptions~\ref{ass:Lipschitz} and \ref{ass:estimation}, it obeys 
\begin{align}
	\mathcal{W}_1\big(f_T(X_T), X_1) < C_1\frac{L_f^3d^{5/2}\log^{5} T}{T} + \varepsilon + \varepsilon_{\mathcal{F}}
\end{align}
	for some universal constant $C_{1} > 0$, where $X_T\sim \mathcal{N}(0,I_d)$. 
% \yuting{is this a high probability statement?}
\end{theorem}

Theorem~\ref{thm:main} implies that, in order to achieve a sampling error of $2(\varepsilon+\varepsilon_{\mathcal{F}})$---in the sense that $\mathcal{W}_1\big(f_T(X_T), X_1) \leq 2 (\varepsilon+\varepsilon_{\mathcal{F}})$---it is sufficient for the number of steps in consistency training to exceed
\begin{align}
	\widetilde{O} \Big(\frac{L_f^3d^{5/2}}{\varepsilon+\varepsilon_{\mathcal{F}}}\Big),
\end{align}
where $d$ denotes the dimension of the target distribution. 
In particular, if the function class $\mathcal{F}$ is rich enough and the approximation error $\varepsilon_{\mathcal{F}}$ equals zero, 
then the number of steps required is about the order of $\frac{L_f^3d^{5/2}}{\varepsilon}.$
This result offers an explicit characterization on the dependence of the Lipschitz constant as well as the dimension of the problem. 
As far as we are aware, this is the first result that theoretically measures the sampling fidelity of consistency models, which serve as a theoretical justification for consistency models as a family of generative models.
% If we compare this result with the theoretical results developed for diffusion models, the best non-asymptotic result developed for probability flow ODE approach satisifes 
% \begin{align*}
% 	TV(q_1, p_1) \lesssim \frac{}
% \end{align*}
As alluded to previously, compared to the popular diffusion models, consistency models bear the benefit of one-step sampling, requiring only a single function evaluation at the sampling stage instead of undergoing recursive denoising. Consequently, our theoretical result provides insights into when one-step sampling is reliable.

We point out that prior results concerning convergence guarantees for diffusion models in terms of the Wasserstein metric often encounter an exponential dependence on the smoothness parameter of the score function (e.g.~\cite{benton2023error,tang2024contractive}). This is mainly due to a direct use of Gr\"{o}nwall's inequality, which provides comparisons to the solution to the initial value problem. 
Tackling this exponential dependence is regarded as a challenging open problem. 
Our result is, however, not directly comparable with these results as the smoothness assumption is imposed instead on the mapping between random variables along the forward trajectory.

Before concluding, 
let us take a moment to provide a brief proof outline for this result; the full technical details are postponed to Sections~\ref{sec:main-proof} and \ref{sec:anuxiliary}.
In order to prove Theorem~\ref{thm:main}, we find it helpful to study how the error   
$\|f_t(X_t)-\Phi_t(X_t)\|_2$ propagates along the probability flow ODE path. 
% To illustrate our main idea, for simplicity, let us ignore the approximation error for now and let $\varepsilon_{\mathcal{F}} = 0.$
Specifically, we establish the following recursive relation for each $t$, where 
\begin{align}
	&\Exs\big[\|f_t(X_t)-\Phi_t(X_t)\|_2\big]  - \mathbb{E}\big[\|f_{t-1}(X_{t-1})-\Phi_{t-1}(X_{t-1})\|_2\big] \notag \\
	\leq &~\mathbb{E}\big[\|f_t(X_t) - f_t^{\mathcal{F}}(X_t)\|_2\big] + \mathbb{E}\big[\|f_t^{\mathcal{F}}(X_t) - f_t^{\star}(X_t)\|_2\big]
	+ \mathbb{E}\Big[\big\|\frac{\partial \Phi_{t-1}}{\partial x}\big(\phi_t(X_{t})\big)\big(\mathbb{E}\big[X_{t-1}\mymid X_t\big] - \phi_t(X_{t})\big)\big\|_2\Big]  \notag\\
	&\qquad\qquad\qquad\qquad\qquad\qquad +\mathbb{E}\Big[\int_0^1 \bigg(\frac{\partial \Phi_{t-1}}{\partial x}(X_{t-1}(\gamma)) - \frac{\partial \Phi_{t-1}}{\partial x}\big(\phi_t(X_{t})\big)\bigg) \big(X_{t-1} - \phi_t(X_{t})\big)\mathrm{d}\gamma\Big]. \label{eqn:intuition}
\end{align}
Here, we denote $X_{t-1}(\gamma) := \gamma X_{t-1} + (1-\gamma) \phi_t(X_{t})$.
If the right-hand side of \eqref{eqn:intuition} can be properly controlled, then Theorem~\ref{thm:main} can be easily established by applying this relation recursively. 
Consequently, it boils down to bounding each term on the right-hand side separately. 
Towards this, the first two terms are concerned with the optimization error and approximation error in training the consistency function, which can be controlled in view of Assumption~\ref{ass:estimation}.
When it comes to the last two terms, while the Lipschitz property of $\Phi_{t\to k}$ allows us to control terms involving derivatives, the main difficulty lies in controlling $\mathbb{E}[X_{t-1}\mymid X_t] - \phi_t(X_{t})$
as well as $X_{t-1} - \phi_t(X_{t}).$ 
Accomplishing this requires a careful study of the probability flow ODE in \eqref{eq:ODE}, the details of which are deferred to Sections~\ref{sec:main-proof} and \ref{sec:anuxiliary}. We would also like to point out that the studies of probability flow ODE are inspired by the framework established in \citet{li2023towards}.

%% file: proof.tex
\section{Proof of Theorem~\ref{thm:main}}
\label{sec:main-proof}

\subsection{Preliminary properties}

Before diving into our main analysis, we collect several auxiliary facts and properties that shall be used frequently throughout this proof. 

\paragraph{Properties of learning rates.}
First, we enumerate some of useful properties about the learning rates as specified by $\{\alpha_t\}$ in \eqref{eqn:alpha-t}.
\begin{subequations}
\label{eqn:properties-alpha-proof}
\begin{align}
	\alpha_t &\geq1-\frac{c_{1}\log T}{T}  \ge \frac{1}{2},\qquad\qquad\quad~~ 1\leq t\leq T \label{eqn:properties-alpha-proof-00}\\
	\frac{1}{2}\frac{1-\alpha_{t}}{1-\overline{\alpha}_{t}} \leq \frac{1}{2}\frac{1-\alpha_{t}}{\alpha_t-\overline{\alpha}_{t}}
	&\leq \frac{1-\alpha_{t}}{1-\overline{\alpha}_{t-1}}  \le \frac{4c_1\log T}{T},\qquad\quad~~ 2\leq t\leq T  \label{eqn:properties-alpha-proof-1}\\
	1&\leq\frac{1-\overline{\alpha}_{t}}{1-\overline{\alpha}_{t-1}} \leq1+\frac{4c_{1}\log T}{T} ,\qquad2\leq t\leq T  \label{eqn:properties-alpha-proof-3} \\
	\overline{\alpha}_{T} & \le \frac{1}{T^{c_2}}. \label{eqn:properties-alpha-proof-alphaT} 
\end{align}
\end{subequations}
In the last line, $c_2 \ge 1000$ is some large numerical constant. All the properties hold provided that $T$ is large enough. The proof of these properties can be found in~\citet[Appendix A.2]{li2023towards}

\paragraph{Truncation on typical events.}
Next, let us introduce the following event:
\begin{align}\label{eqn:eset}
	\mathcal{E}_t \defn \bigg\{(x_t, x_{t-1}) \in \mathbb{R}^d \times \mathbb{R}^d ~\Big| -\log p_{X_t}(x_t) \leq c_3 d\log T, ~\|x_{t-1} - x_t/\sqrt{\alpha_t}\|_2 \leq c_4 \sqrt{d(1 - \alpha_t)\log T} \bigg\},
\end{align}
where $c_3$ and $c_4$ are some numerical constants to be specified later.
Generally speaking, $\mathcal{E}$ encompasses a typical range of the values of $(X_t,X_{t-1})$, and 
some part of our analysis proceed by seperately considering the points in $\mathcal{E}$ and those outside $\mathcal{E}$. 
While truncated on $\mathcal{E}$, there are some nice continuity properties on the trajectories, and for $(x_t,x_{t-1}) \in \mathcal{E}^c$, we have
\begin{align}
	\mathbb{P}\big((X_{t},X_{t-1})\notin\mathcal{E}\big) & =\int_{(x_{t},x_{t-1})\notin\mathcal{E}}p_{X_{t-1}}(x_{t-1})p_{X_{t}\mymid X_{t-1}}(x_{t}\mymid x_{t-1})\mathrm{d}x_{t-1}\mathrm{d}x_{t} \notag\\
	 & =\int_{(x_{t},x_{t-1})\notin\mathcal{E}}p_{X_{t}-1}(x_{t-1})\frac{1}{\big(2\pi(1-\alpha_{t})\big)^{d/2}}\exp\bigg(-\frac{\|x_{t}-\sqrt{\alpha_{t}}x_{t-1}\|_{2}^{2}}{2(1-\alpha_{t})}\bigg)\mathrm{d}x_{t-1}\mathrm{d}x_{t} \notag\\
	 & \le\exp\big(-c_4d\log T\big),
		\label{eq:P-Xt-X-t-1-not-E}
\end{align}
which can be a high order term in $T$ when $c_4$ is large enough. 

On the typical event $\mathcal{E}$, the score and density functions behave regularly, which are clarified by the following two lemmas from \cite{li2023towards}. 
\begin{lemma}[\citet{li2023towards}, Lemma 1] 
	\label{lemma:x0}
	Consider any $x_t \in \mathbb{R}^d$ satisfying $-\log p_{X_t}(x_t) \leq c_3 d\log T$
	for some large enough constant $c_3$. Then
	it holds that
	\begin{subequations}
	\begin{align}
		\mathbb{E}\left[\big\| \sqrt{\overline{\alpha}_{t}}X_{0} - x_t \big\|_{2}\,\big|\,X_{t}=x_t\right] &\lesssim \sqrt{d(1-\overline{\alpha}_{t})\log T},\label{eq:E-xt-X0} \\
		\mathbb{E}\left[\big\| \sqrt{\overline{\alpha}_{t}}X_{0} - x_t \big\|^2_{2}\,\big|\,X_{t}=x_t\right] &\lesssim  d(1-\overline{\alpha}_{t})\log T,\label{eq:E2-xt-X0} \\
		\mathbb{E}\left[\big\| \sqrt{\overline{\alpha}_{t}}X_{0} - x_t \big\|^3_{2}\,\big|\,X_{t}=x_t\right] &\lesssim \big(d(1-\overline{\alpha}_{t})\log T\big)^{3/2}.\label{eq:E3-xt-X0}
	\end{align}
	\end{subequations}
\end{lemma}
\noindent
Lemma~\ref{lemma:x0} implies that if $X_t$ taking on a ``typical'' value, then condition on it, 
the vector $\sqrt{\overline{\alpha}_{t}}X_{0} - X_t  = \sqrt{1-\overline{\alpha}_t} \,\overline{W}_t$ might still follow a sub-Gaussian tail, 
whose expected norm remains on the same order of that of an unconditional Gaussian vector $\mathcal{N}(0, (1-\overline{\alpha}_t)I_d)$. 

\begin{lemma}[\citet{li2023towards}, Lemma 2]
	\label{lemma:river}
	Consider any two points $x_t, x_{t-1}\in \mathbb{R}^d$ obeying
\begin{align}
	-\log p_{X_t}(x_t) \leq \frac{1}{2} c_3 d\log T, 
	\quad\text{and}\quad \bigg\|x_{t-1} -   \frac{x_t}{ \sqrt{\alpha_t}} \bigg\|_2 \leq c_4 \sqrt{d(1 - \alpha_t)\log T}
	\label{eq:assumption-lem:river}
\end{align}
for some large constants $c_3, c_4>0$. 
Then we have 
\begin{align*}
	p_{X_{t-1}}(x) = \bigg(1+O\Big(\sqrt{\frac{d(1-\alpha_t)\log T}{1-\overline{\alpha}_t}}\Big)\bigg)p_{X_t}(x),
\end{align*}
and for all $\gamma \in [0,1]$,
\begin{align}
\label{eqn:river}
	-\log p_{X_{t-1}}\big(x_{t}(\gamma)\big) & \leq  c_6 d\log T.
\end{align}

\end{lemma}
In other words, Lemma~\ref{lemma:river} ensures that if $x_t$ falls within a typical set of $X_t$ and the point $x_{t-1}$ is not too far away from $x_t/\sqrt{\alpha_t}$, 
then $x_{t-1}$ is also a typical value of $X_{t-1}$. 
Lemma~\ref{lemma:river} here is in a slightly different form from the original version in \cite{li2023towards} due to a different definition of $x_{t-1}(\gamma)$. 
Notice that using the inequality~\eqref{eq:lemma2-con}, the proof of Lemma 2 in \cite{li2023towards} remains valid with the new definition of $x_{t-1}(\gamma)$, so we keep the original statement of this lemma.

% \yuting{to be rewritten/checked}

\subsection{Main analysis}

Throughout this proof, we shall use capital letters to denote random vectors, and lower case letters to denote their corresponding realizations, i.e. for some specific point in the sample space $\omega \in \Omega$, we could write $x_t := X_t(\omega)$ and $x_{t-1} := X_{t-1}(\omega)$.

First, notice that $X_1 \overset{\mathsf{d}}{=} \Phi_{T}(X_T)$,
which gives 
\begin{align*}
\mathcal{W}_1\big(f_T(X_T), X_1) = \mathcal{W}_1\big(f_T(X_T), \Phi_{T}(X_T)\big)
\le \mathbb{E}\big[\|f_T(X_T) - \Phi_{T}(X_T)\|_2\big].
\end{align*}
To control the right hand side above, let us introduce a piece of notation  
\begin{align}
	\xi_t(x) &:= f_t(x) - \Phi_t(x),
\end{align}
and we claim that $\xi_t$ satisfies the following recursive relation with $\xi_1 = 0$:
\begin{align}
\label{eq:recursion}
\xi_t(x_t) &= \mathbb{E}\big[\xi_{t-1}(X_{t-1})\mymid X_t = x_t\big] + \big(f_t(x_t) - f_t^{\mathcal{F}}(x_t)\big) + \big(f_t^{\mathcal{F}}(x_t) - f_t^{\star}(x_t)\big) \notag \\ 
&\qquad+ \frac{\partial \Phi_{t-1}}{\partial x}\big(\phi_t(x_{t})\big)\big(\mathbb{E}\big[X_{t-1}\mymid X_t = x_t\big] - \phi_t(x_{t})\big) \notag\\
&\qquad\qquad+ \mathbb{E}\bigg[\int_0^1 \bigg(\frac{\partial \Phi_{t-1}}{\partial x}(X_{t-1}(\gamma)) - \frac{\partial \Phi_{t-1}}{\partial x}\big(\phi_t(x_{t})\big)\bigg) \big(X_{t-1} - \phi_t(x_{t})\big)\mathrm{d}\gamma\,\Big|\, X_t = x_t\bigg],
\end{align}
where we let 
\begin{align}
	x_{t-1}(\gamma) := \gamma x_{t-1} + (1-\gamma) \phi_t(x_{t})
\end{align}
for $\gamma \in [0,1]$.
We leave its derivation to Section~\ref{sec:pf-recursion}.
In addition, let us denote $X_{t-1}(\gamma) = \gamma X_{t-1} + (1-\gamma) \phi_t(X_{t})$, and the above relation implies that
\begin{align}
\mathbb{E}\big[\big\|\xi_T(X_T)\big\|_2\big] &\le \mathbb{E}\big[\big\|\xi_{T-1}(X_{T-1})\big\|_2\big] + \mathbb{E}\big[\big\|f_T(X_T) - f_T^{\mathcal{F}}(X_T)\big\|_2\big] + \mathbb{E}\big[\big\|f_T^{\mathcal{F}}(X_T) - f_T^{\star}(X_T)\big\|_2\big]\notag\\ 
&\qquad~~ + \mathbb{E}\bigg\{\bigg\|\frac{\partial \Phi_{T-1}}{\partial x}\big(\phi_T(X_{T})\big)\big(\mathbb{E}\big[X_{T-1}\mymid X_T\big] - \phi_T(X_{T})\big)\bigg\|_2 \notag\\
&\qquad\qquad~ + \mathbb{E}\bigg[\bigg\|\int_{0}^1 \bigg(\frac{\partial \Phi_{T-1}}{\partial x}(X_{T-1}(\gamma)) - \frac{\partial \Phi_{T-1}}{\partial x}\big(\phi_T(X_{T})\big)\bigg) \big(X_{T-1} - \phi_T(X_{T})\big)\mathrm{d}\gamma\bigg\|_2\bigg]\bigg\} \label{eqn:brahms}\\
&\stackrel{(\text{i})}{\le} \sum_{t=1}^T \mathbb{E}\big[\big\|f_t(X_t) - f_t^{\mathcal{F}}(X_t )\big\|_2\big] + \mathbb{E}\big[\big\|f_t^{\mathcal{F}}(X_t) - f_t^{\star}(X_t )\big\|_2\big] \notag \\
&\qquad\qquad+ \sum_{t = 2}^{T} \bigg\{\frac{\partial \Phi_{t-1}}{\partial x}\big(\phi_t(X_{t})\big)\mathbb{E}\Big[\big\|\mathbb{E}\big[X_{t-1}\mymid X_t\big] - \phi_t(X_{t})\big\|_2\Big] \notag\\
&\qquad\qquad\qquad\quad~~+ \int_0^1\mathbb{E}\bigg[\bigg\|\frac{\partial \Phi_{t-1}}{\partial x}(X_{t-1}(\gamma)) - \frac{\partial \Phi_{t-1}}{\partial x}\big(\phi_t(X_{t})\big)\bigg\| \big\|X_{t-1} - \phi_t(X_{t})\big\|_2\mathrm{d}\lambda\bigg]\bigg\}\notag\\
&\stackrel{(\text{ii})}{\le} \varepsilon + \varepsilon_{\mathcal{F}} + \sum_{t = 2}^{T} \bigg\{L_f\mathbb{E}\Big[\big\|\mathbb{E}\big[X_{t-1}\mymid X_t\big] - \phi_t(X_{t})\big\|_2\Big] \notag\\
&\qquad+ \sup_{\gamma} \mathbb{E}\bigg[\bigg\|\frac{\partial \Phi_{t-1}}{\partial x}(X_{t-1}(\gamma)) - \frac{\partial \Phi_{t-1}}{\partial x}\big(\phi_t(X_{t})\big)\bigg\| \big\|X_{t-1} - \phi_t(X_{t})\big\|_2\bigg]\bigg\}, \notag \\
&= \varepsilon + \varepsilon_{\mathcal{F}} + T_1 + T_2, \label{eq:main}
\end{align}
where relation $(\text{i})$ applies inequality~\eqref{eqn:brahms} recursively
and relation $(\text{ii})$ invokes the triangle inequality and Assumption~\ref{ass:Lipschitz}.
In the following, we proceed to bound the latter two terms separately.

\paragraph{Control quantity $T_1$.}
Let us start with the term $T_{1}$, where the goal is to control each quantity in the summation, which is 
\mbox{$\mathbb{E}\Big[\big\|\mathbb{E}\big[X_{t-1}\mymid X_t\big]-\phi_t(X_t) \big\|_2^2\Big]$}. 
Recalling the backward ODE flow~\eqref{eq:ODE} that $\Phi_{t \to k}(x) \defn g_t(x,\overline{\alpha}_k)$ and 
\begin{align*}
\frac{\partial g_t(x,\overline{\alpha})}{\partial \overline{\alpha}} 
= \frac{1}{2\overline{\alpha}} \Big(g_t(x,\overline{\alpha}) + s_{\overline{\alpha}}(g_t(x,\overline{\alpha}))\Big),
\qquad\text{and }g_t(x,\overline{\alpha}_t) = x,
\end{align*}
it is easy to check that 
\begin{align}
\label{eq:tranformed ODE}
	\frac{\partial\Big(\frac{1}{\sqrt{\overline{\alpha}}}g_t(x,\overline{\alpha})\Big)}{\partial \overline{\alpha}} = \frac{1}{2\overline{\alpha}^{\frac{3}{2}}}s_{\overline{\alpha}}(g_t(x,\overline{\alpha})).
\end{align}
As a result, we can track the backward process with the score function as:
\begin{align}
\sqrt{\alpha_t}\phi_t(X_t) &= X_t + (\sqrt{\alpha_t}\Phi_{t\to t-1}(X_t) - \Phi_{t\to t}(X_t))\notag\\
&= X_t + \sqrt{\overline{\alpha}_t} \int_{\overline{\alpha}_t}^{\overline{\alpha}_{t-1}}
\frac{1}{2\overline{\alpha}^{3/2}}s_{\overline{\alpha}}(g_t(X_t,\overline{\alpha}))\mathrm{d}\overline{\alpha} \notag\\
&= X_t + (1 - \sqrt{\alpha_t})s_t(X_t) + \frac{1}{2}\int_{\overline{\alpha}_t}^{\overline{\alpha}_{t-1}}\sqrt{\frac{\overline{\alpha}_t}{\overline{\alpha}^3}}\big(s_{\overline{\alpha}}(g_t(X_t,\overline{\alpha})) - s_t(X_t)\big)\mathrm{d}\overline{\alpha}.\label{eq:phi-t-x}
\end{align}
For the remaining term, we first apply the definition of the forward process:
\begin{align}
\sqrt{\alpha_t}\mathbb{E}\big[X_{t-1}\mymid X_t\big] 
&= \sqrt{\alpha_t}\mathbb{E}\Big[\sqrt{\overline{\alpha}_{t-1}}X_0 + \sqrt{1-\overline{\alpha}_{t-1}}Z\mymid X_t = \sqrt{\overline{\alpha}_{t}}X_0 + \sqrt{1-\overline{\alpha}_{t}}Z \Big] \notag\\
&= X_t + \mathbb{E}\big[\big(\sqrt{\alpha_{t}-\overline{\alpha}_{t}} - \sqrt{1-\overline{\alpha}_{t}}\big)Z\mymid X_t = \sqrt{\overline{\alpha}_{t}}X_0 + \sqrt{1-\overline{\alpha}_{t}}Z\big].\label{eq:conditional expectation}
\end{align}
The previous work on score matching admits a minimum mean square error (MMSE) form for the score function
(e.g.~\cite{hyvarinen2005estimation,vincent2011connection,chen2022sampling}):
\begin{align*}
	s_{\overline{\alpha}} := \arg\min_{s:\mathbb{R}^d \to \mathbb{R}^d} \mathbb{E}\bigg[\bigg\|s(\sqrt{\overline{\alpha}}X_{0}+\sqrt{1-\overline{\alpha}}Z)+\frac{1}{\sqrt{1-\overline{\alpha}}}Z\Big\|_2^2\bigg],
\end{align*}
which leads to an alternative expression by the change of variables:
\begin{align}
	s_{\overline{\alpha}}(x) = \mathbb{E}\bigg[-\frac{1}{\sqrt{1-\overline{\alpha}}}Z~\Big|~\sqrt{\overline{\alpha}}X_{0}+\sqrt{1-\overline{\alpha}}Z = x\bigg].\label{eq:msse}
\end{align}
Plugging equation~\eqref{eq:msse} into \eqref{eq:conditional expectation}, we obtain
\begin{align}
	\sqrt{\alpha_t}\mathbb{E}\big[X_{t-1}\mymid X_t\big] = X_t + \big(1-\overline{\alpha}_{t} - \sqrt{(1-\overline{\alpha}_{t})(\alpha_{t}-\overline{\alpha}_{t})}\big)s_t(X_{t}),\label{eq:conditional-expectation-final}
\end{align}
which when combined with \eqref{eq:conditional expectation} yields
\begin{align}
\label{eqn:beethoven}
\notag &\sqrt{\alpha_t}\phi_t(X_t) - \sqrt{\alpha_t}\mathbb{E}\big[X_{t-1}\mymid X_t\big]\\
&~=
\Big((1 - \sqrt{\alpha_t}) - (1-\overline{\alpha}_{t}) + \sqrt{(1-\overline{\alpha}_{t})(\alpha_{t}-\overline{\alpha}_{t})}\Big)s_t(X_{t})
+ \frac{1}{2}\int_{\overline{\alpha}_t}^{\overline{\alpha}_{t-1}}\sqrt{\frac{\overline{\alpha}_t}{\overline{\alpha}^3}}\big(s_{\overline{\alpha}}(g_t(X_t,\overline{\alpha})) - s_t(X_{t})\big)\mathrm{d}\overline{\alpha}.
\end{align}

With equations~\eqref{eq:phi-t-x} and \eqref{eq:conditional-expectation-final} in place, we arrive at
\begin{align}
	\mathbb{E}\Big[\big\|\phi_t(X_t) - \mathbb{E}\big[X_{t-1}\mymid X_t\big]\big\|_2^2\Big]
	&= \frac{1}{\alpha_t} \mathbb{E}\Big[\big\|\sqrt{\alpha_t}\phi_t(X_t) - \sqrt{\alpha_t}\mathbb{E}\big[X_{t-1}\mymid X_t\big]\big\|_2^2\Big]\notag\\
	&= \frac{1}{\alpha_t}\mathbb{E}\Bigg[\Big\|\Big( - (1-\overline{\alpha}_{t}) + \sqrt{(1-\overline{\alpha}_{t})(\alpha_{t}-\overline{\alpha}_{t})} + (1 - \sqrt{\alpha_t})\Big)s_t(X_{t})\notag\\
	&\qquad\qquad\qquad\qquad\quad + \frac{1}{2}\int_{\overline{\alpha}_t}^{\overline{\alpha}_{t-1}}\sqrt{\frac{\overline{\alpha}_t}{\overline{\alpha}^3}}\big(s_{\overline{\alpha}}(g_t(X_t,\overline{\alpha})) - s_t(X_{t})\big)\mathrm{d}\overline{\alpha}\Big\|_2^2 \Bigg]\notag\\
	&\le \frac{2}{\alpha_t}\Big(-\sqrt{1-\overline{\alpha}_{t}}~\big(\sqrt{1-\overline{\alpha}_{t}}-\sqrt{\alpha_{t}-\overline{\alpha}_{t}}\big)+ (1-\sqrt{\alpha_t})\Big)^2\mathbb{E}\big[\|s_t(X_{t})\|_2^2\big] \notag\\
	&\qquad\qquad\qquad+\frac{1}{2\alpha_t}\mathbb{E}\Big[\Big\|\int_{\overline{\alpha}_t}^{\overline{\alpha}_{t-1}}\sqrt{\frac{\overline{\alpha}_t}{\overline{\alpha}^3}}\big(s_{\overline{\alpha}}(g_t(X_t,\overline{\alpha})) - s_t(X_{t})\big)\mathrm{d}\overline{\alpha}\Big\|_2^2\Big].\notag
\end{align}
In view of the Taylor expansion, we can further control the right hand side above as 
\begin{align}
\mathbb{E}\Big[\big\|\phi_t(X_t) - \mathbb{E}\big[X_{t-1}\mymid X_t\big]\big\|_2^2\Big]
	&\le \frac{2}{\alpha_t}\bigg(-\Big(\frac{1-\alpha_t}{2} - \frac{(1-\alpha_t)^2}{8(1-\overline{\alpha}_t)}\Big)+\Big(\frac{1-\alpha_t}{2} + \frac{(1-\alpha_t)^2}{8}\Big)\bigg)^2\mathbb{E}\big[\|s_t(X_{t})\|_2^2\big] \notag\\
	&\qquad\qquad + O\Big(\frac{(1-\alpha_t)^5}{\alpha_t(1-\overline{\alpha}_t)^{5/2}}\Big)\mathbb{E}\big[\|s_t(X_{t})\|_2^2\big]\notag\\
	&\qquad\qquad\qquad+\frac{1}{2\alpha_t}\mathbb{E}\Big[\Big\|\int_{\overline{\alpha}_t}^{\overline{\alpha}_{t-1}}\sqrt{\frac{\overline{\alpha}_t}{\overline{\alpha}^3}}\big(s_{\overline{\alpha}}(g_t(X_t,\overline{\alpha})) - s_t(X_{t})\big)\mathrm{d}\overline{\alpha}\Big\|_2^2\Big]\notag\\
	&\lesssim \frac{(1-\alpha_t)^4}{(1-\overline{\alpha}_t)^2}\mathbb{E}\big[\|s_t(X_{t})\|_2^2\big] + \mathbb{E}\Big[\Big\|\int_{\overline{\alpha}_t}^{\overline{\alpha}_{t-1}}\sqrt{\frac{\overline{\alpha}_t}{\overline{\alpha}^3}}\big(s_{\overline{\alpha}}(g_t(X_t,\overline{\alpha})) - s_t(X_{t})\big)\mathrm{d}\overline{\alpha}\Big\|_2^2\Big].\label{eq: phi - E}
\end{align}

To further control the right hand side of expression~\eqref{eq: phi - E}, we introduce the following Lemma~\ref{lemma:score moment} and Lemma~\ref{lemma:discrete}, which provide upper bounds for the two expectations in \eqref{eq: phi - E} respectively. 
The proofs of these lemmas can be found in Sections~\ref{sec:pf-lem-score-2} and \ref{sec:pf-lem-discrete} respectively. 

\begin{lemma}\label{lemma:score moment}
	For $X_t \sim \sqrt{\overline{\alpha}_{t}}X_0 + \sqrt{1-\overline{\alpha}_{t}}\ Z$, where  $X_0 \sim \Pdata$ and $Z \sim \mathcal{N}(0, I_d)$, the second moment of the score function satisfies
\begin{align*}
	\mathbb{E}\big[\|s_t(X_{t})\|_2^2\big] \le \frac{d}{1-\overline{\alpha}_{t}}.
\end{align*}
Moreover, for any $0 < \overline{\alpha} < 1$, the lemma still holds when replace $\overline{\alpha}_t$ with $\overline{\alpha}$.
\end{lemma}
\begin{lemma}\label{lemma:discrete}
	For $X_t$ defined the same as in Lemma~\ref{lemma:score moment}, pre-selected $\{\alpha_i\}_{1\le i\le t}$ and corresponding $\overline{\alpha}_t$, $\overline{\alpha}_{t-1}$, we deduce that
\begin{align} 
\mathbb{E}\Big[\Big\|\int_{\overline{\alpha}_t}^{\overline{\alpha}_{t-1}}\sqrt{\frac{\overline{\alpha}_t}{\overline{\alpha}^3}}\big(s_{\overline{\alpha}}(g_t(X_t,\overline{\alpha})) - s_t(X_{t})\big)\mathrm{d}\overline{\alpha}\Big\|_2^2\Big] \lesssim \frac{(1 - \alpha_t)^4d^3\log^3 T}{(1-\overline{\alpha}_{t})^3}\notag.
\end{align}
\end{lemma}
In view of Lemma~\ref{lemma:score moment} and Lemma~\ref{lemma:discrete}, the right hand side of \eqref{eq: phi - E} is further controlled as
\begin{align}
	\mathbb{E}\Big[\big\|\phi_t(X_t) - \mathbb{E}\big[X_{t-1}\mymid X_t\big]\big\|_2^2\Big]
	&\lesssim \frac{(1-\alpha_t)^4 d}{(1-\overline{\alpha}_{t})^3} +\frac{(1 - \alpha_t)^4d^3\log^3 T}{(1-\overline{\alpha}_{t})^3}\notag\\
	&\lesssim \frac{(1 - \alpha_t)^4d^3\log^3 T}{(1-\overline{\alpha}_{t})^3}.
\end{align}
Now by properties of the step sizes mentioned in \eqref{eqn:properties-alpha-proof-1}, this upper bound can be simplified as 
\begin{align}
	\mathbb{E}\Big[\big\|\phi_t(X_t) - \mathbb{E}\big[X_{t-1}\mymid X_t\big]\big\|_2\Big]
	\le \frac{C_2d^{3/2}\log^{7/2}T}{T^2},\label{eq;first}
\end{align}
where $C_2$ denotes some universal constant.

\paragraph{Control quantity $T_2$.} Now, let us turn our attention to control the term $T_{2}$.
We first decompose this term by the Cauchy-Schwartz inequality:
\begin{align}
	&\mathbb{E}\bigg[\bigg\|\frac{\partial \Phi_{t-1}}{\partial x}(X_{t-1}(\gamma)) - \frac{\partial \Phi_{t-1}}{\partial x}\big(\phi_t(X_{t})\big)\bigg\| \big\|X_{t-1} - \phi_t(X_{t})\big\|_2\bigg]\notag\\
	\le& \frac{1}{2}\mathbb{E}\big\|X_{t-1} - \phi_t(X_{t})\big\|_2^2 + \frac{1}{2}\mathbb{E}\bigg\|\frac{\partial \Phi_{t-1}}{\partial x}(X_{t-1}(\gamma)) - \frac{\partial \Phi_{t-1}}{\partial x}\big(\phi_t(X_{t})\big)\bigg\|^2,\label{eq:cs}
\end{align}
and we aim to handle the two components respectively.

\begin{itemize}
\item Towards bounding the first term in \eqref{eq:cs}, in view of relation~\eqref{eq:phi-t-x}, we make the observation that
\begin{align}
\mathbb{E}\big\|X_{t-1} - \phi_t(X_{t})\big\|_2^2
&=\frac{1}{\alpha_t}\mathbb{E}\big\|({\sqrt{\alpha_t}}X_{t-1}-X_t) - ({\sqrt{\alpha_t}}\phi_t(X_t)-X_t)\big\|_2^2\notag\\
&=\frac{1}{\alpha_t}\mathbb{E}\Big\|\big(\sqrt{\alpha_{t}-\overline{\alpha}_{t}} - \sqrt{1-\overline{\alpha}_{t}}\big)Z + (1 - \sqrt{\alpha_t})s_t(X_t)\notag\\
&\qquad\qquad\qquad\qquad\qquad\qquad\quad+ \frac{1}{2}\int_{\overline{\alpha}_t}^{\overline{\alpha}_{t-1}}\sqrt{\frac{\overline{\alpha}_t}{\overline{\alpha}^3}}\big(s_{\overline{\alpha}}(g_t(X_t,\overline{\alpha})) - s_t(X_t)\big)\mathrm{d}\overline{\alpha}\Big\|_2^2\notag\\
&\le \frac{3}{\alpha_t}\mathbb{E}\bigg[\big\|\big(\sqrt{\alpha_{t}-\overline{\alpha}_{t}} - \sqrt{1-\overline{\alpha}_{t}}\big)Z\big\|_2^2 + \big\|(1 - \sqrt{\alpha_t})s_t(X_{t})\big\|_2^2\notag\\ 
&\qquad\qquad\qquad\qquad\qquad\qquad+ \frac{1}{4}\Big\|\int_{\overline{\alpha}_t}^{\overline{\alpha}_{t-1}}\sqrt{\frac{\overline{\alpha}_t}{\overline{\alpha}^3}}\big(s_{\overline{\alpha}}(g_t(X_t,\overline{\alpha})) - s_t(X_{t})\big)\mathrm{d}\overline{\alpha}\Big\|_2^2\bigg] \notag\\
&\lesssim \frac{(1-\alpha_t)^2}{1-\overline{\alpha}_t} + \frac{(1-\alpha_{t})^2d}{1-\overline{\alpha}_{t}} + \frac{(1 - \alpha_t)^4d^3\log^3 T}{(1-\overline{\alpha}_{t})^3}\notag\\ 
&\lesssim \frac{d \log^2 T}{T^2}.\label{eq:second-term-1}
\end{align}
Here, we recall the properties of the learning rates as in \eqref{eqn:properties-alpha-proof-00} and \eqref{eqn:properties-alpha-proof-1}. 

\item When it comes to the second term in \eqref{eq:cs} , we claim that for any $(x_t,x_{t-1})$ pair, it can be decomposed as
\begin{align} \label{eq:Jacobi-all}
\bigg\|\frac{\partial \Phi_{t-1}}{\partial x}(x_{t-1}(\gamma)) - \frac{\partial \Phi_{t-1}}{\partial x}\big(\phi_t(x_{t})\big)\bigg\| \le L_f^2\sum_{k = 1}^{t} \bigg\|\frac{\partial \phi_{k}}{\partial x}\big(\Phi_{t-1 \to k}(x_{t-1}(\gamma))\big) - \frac{\partial \phi_{k}}{\partial x}\big(\Phi_{t \to k}(x_t)\big)\bigg\|.
\end{align}
The proof of claim~\eqref{eq:Jacobi-all} is provided in our Section~\ref{sec:pf-jacobi}. 
We proceed to control the right hand side above with the aid of the following lemma. 
\begin{lemma}
\label{lemma:Jacobi-diff}
For $2 \le k < t \le T$, $X_t$ and $X_{t-1}(\gamma)$ defined as above, it holds that 
\begin{align} 
\mathbb{E}\bigg\|\frac{\partial \phi_{k}}{\partial x}\big(\Phi_{t-1 \to k}(X_{t-1}(\gamma))\big) - \frac{\partial \phi_{k}}{\partial x}\big(\Phi_{t \to k}(X_t)\big)\bigg\|^2\!\! \lesssim\! \frac{(1-\alpha_{k})^2(1-\alpha_t)^2L_f^2d^4\log^3 T}{(1-\overline{\alpha}_{k})^2(1 - \overline{\alpha}_t)^2} + \frac{(1-\alpha_t)^4d^4\log^4 T}{(1 - \overline{\alpha}_t)^4}.
\end{align}
\end{lemma}
We defer the proof of this result to Section~\ref{sec:pf-jacobi}. 
With Lemma~\ref{lemma:Jacobi-diff} in place, we can further derive that
\begin{align}
	\mathbb{E}\bigg\|\frac{\partial \Phi_{t-1}}{\partial x}(x_{t-1}(\gamma)) - \frac{\partial \Phi_{t-1}}{\partial x}\big(\phi_t(x_{t})\big)\bigg\|^2 
	&\lesssim L_f^4 T^2 \bigg(\frac{(1-\alpha_k)^2(1-\alpha_t)^2L_f^2d^4\log^3 T}{(1-\overline{\alpha}_k)^2(1-\overline{\alpha}_t)^2} + \frac{(1-\alpha_t)^4d^4 \log^4 T}{(1-\overline{\alpha}_t)^4}\bigg)\notag\\
	&\lesssim \frac{L_f^6 d^4\log^8 T}{T^2}.\label{eq:second-term-2}
\end{align}
Here, again we use the properties of step size in \eqref{eqn:properties-alpha-proof-1}.
\end{itemize}
Putting expressions~\eqref{eq:second-term-1} and \eqref{eq:second-term-2} together leads to
\begin{align}
\mathbb{E}\bigg[\bigg\|\frac{\partial \Phi_{t-1}}{\partial x}(X_{t-1}(\gamma)) - \frac{\partial \Phi_{t-1}}{\partial x}\big(\phi_t(X_{t})\big)\bigg\| \big\|X_{t-1} - \phi_t(X_{t})\big\|_2\bigg] 
&\le \frac{C_1 L_f^3d^{5/2}\log^{5} T}{T^2}.\label{eq:second}
\end{align}

\paragraph{In conclusion,} taking relations~\eqref{eq;first} and \eqref{eq:second} collectively with relation~\eqref{eq:main}, we arrive at  
\begin{align}
\mathcal{W}_1\big(f_T(X_T), X_1) \le \mathbb{E}\big[\big\|f_T(X_T) - \Phi_T(X_T)\big\|_2\big] 
&\le \frac{C_2 d^{3/2}\log^{7/2}T}{T^2} \cdot T + \frac{C_1 L_f^3d^{5/2}\log^{5} T}{T^2} \cdot T + \varepsilon+ \varepsilon_{\mathcal{F}}\notag\\
&\le \frac{C_1L_f^3d^{5/2}\log^{5} T}{T} + \varepsilon + \varepsilon_{\mathcal{F}}.\notag
\end{align}
This thus completes the proof of our advertised result.

%% file: discussion.tex
\section{Discussion}

In this work, we have developed a rigorous mathematical framework for analyzing consistency training in diffusion models.
Given a set of consistency functions with sufficiently small training error, we have pinned down the finite-sample performance for the consistency model in terms of the Wasserstein metric, with explicit dependencies on the problem parameters. 
The analysis framework laid out in the current paper might potentially be applicable to other generative and distillation models, such as the progressive training procedure in \cite{salimans2022progressive}.

Moving forward, we highlight several possible directions worthy of future investigation. 
For instance, it remains unclear whether our theory offers optimal dependencies on the Lipschitz constant of the $\Phi_{t \to k}$ mappings and the ambient dimension $d$. Can we further refine our theory in order to obtain tighter dependencies or establish matching lower bounds? 
In addition, our theory decouples the training phase from the sampling phase by assuming a small optimization/estimation error. 
It would be of great interest to consider whether one can establish end-to-end results that combine these two phases. 
Moving beyond consistency models, it would also be interesting to compare our theory---in terms of sampling efficiency---with other generative sampling methods, such as accelerated ODE and SDE methods~\citep{song2020improved,lu2022dpm}.

%% file: proof-auxiliary.tex
\section{Proof of auxiliary results}
\label{sec:anuxiliary}

% \yuting{use subsection instead of paragraph for each proof}

\subsection{Proof of the recursion~\eqref{eq:recursion}}
\label{sec:pf-recursion}

Recalling the definitions of $f_t(x)$ and $f^{\star}_t(x)$ yields
\begin{align}
f_t(x_t) &= \mathbb{E}\big[\Phi_{t-1}(X_{t-1})\mymid X_t = x_t\big] + \mathbb{E}\big[\xi_{t-1}(X_{t-1})\mymid X_t = x_t\big] + \big(f_t(x_t) - f_t^{\mathcal{F}}(x_t)\big) + \big(f_t^{\mathcal{F}}(x_t) - f_t^{\star}(x_t)\big) \notag\\
&= \Phi_{t}(x_{t}) + \mathbb{E}\big[\Phi_{t-1}(X_{t-1}) - \Phi_{t-1}\big(\phi_t(x_{t})\big)\mymid X_t = x_t\big] + \mathbb{E}\big[\xi_{t-1}(x_{t-1})\mymid X_t = x_t\big] \notag\\
&\qquad\qquad+ \big(f_t(x_t) - f_t^{\mathcal{F}}(x_t)\big) + \big(f_t^{\mathcal{F}}(x_t) - f_t^{\star}(x_t)\big). \notag
\end{align}
Invoking the Taylor expansion to obtain 
\begin{align*}
\Phi_{t-1}(x_{t-1}) - \Phi_{t-1}\big(\phi_t(x_{t})\big)
&= \int_{\gamma} \frac{\partial \Phi_{t-1}}{\partial x}(x(\gamma)) \big(x_{t-1} - \phi_t(x_{t})\big)\mathrm{d}\gamma \\
&= \frac{\partial \Phi_{t-1}}{\partial x}\big(\phi_t(x_{t})\big)\big(x_{t-1} - \phi_t(x_{t})\big) \\
&\qquad\qquad+ \int_{\gamma} \bigg(\frac{\partial \Phi_{t-1}}{\partial x}(x_{t-1}(\gamma)) - \frac{\partial \Phi_{t-1}}{\partial x}\big(\phi_t(x_{t})\big)\bigg) \big(x_{t-1} - \phi_t(x_{t})\big)\mathrm{d}\gamma,
 \end{align*}
further leads to 
\begin{align}
f_t(x_t)&= \Phi_{t}(x_{t}) + \mathbb{E}\big[\xi_{t-1}(X_{t-1})\mymid X_t = x_t\big] + \big(f_t(x_t) - f_t^{\mathcal{F}}(x_t)\big) + \big(f_t^{\mathcal{F}}(x_t) - f_t^{\star}(x_t)\big) \notag\\
&\qquad\qquad+ \frac{\partial \Phi_{t-1}}{\partial x}\big(\phi_t(x_{t})\big)\big(\mathbb{E}\big[X_{t-1}\mymid X_t = x_t\big] - \phi_t(x_{t})\big) \notag\\
&\qquad\qquad\qquad\qquad+ \mathbb{E}\bigg[\int_{0}^1 \bigg(\frac{\partial \Phi_{t-1}}{\partial x}(X_{t-1}(\gamma)) - \frac{\partial \Phi_{t-1}}{\partial x}\big(\phi_t(x_{t})\big)\bigg) \big(X_{t-1} - \phi_t(x_{t})\big)\mathrm{d}\gamma\,\Big|\, X_t = x_t\bigg].
\end{align}
This thus establishes relation~\eqref{eq:recursion}.

\subsection{Proof of Lemma~\ref{lemma:score moment}}
\label{sec:pf-lem-score-2}

We first recall the definition of $s_t(x)$, which is the score function of $X_t = \sqrt{\overline{\alpha}_t}X_0 + \sqrt{1-\overline{\alpha}_{t}}Z$.
If we let $P_{\sqrt{\overline{\alpha}_t}}$ be the probability measure of $\sqrt{\overline{\alpha}_t}X_0$, and $p_{\sqrt{1-\overline{\alpha}_{t}}Z}$ be the density of $\sqrt{1-\overline{\alpha}_{t}}Z$, 
by definition of the score function, we can write  
\begin{align}
	s_t(x) 
	&= -\frac{\nabla_x \int_{x_0}p_{\sqrt{1-\overline{\alpha}_{t}}Z}(x-\sqrt{\overline{\alpha}_t}x_0)\mathrm{d}P_{\sqrt{\overline{\alpha}_t}X_{0}}(\sqrt{\overline{\alpha}_t}x_0)}{\int_{x_0}p_{\sqrt{1-\overline{\alpha}_{t}}Z}(x-\sqrt{\overline{\alpha}_t}x_0)\mathrm{d}P_{\sqrt{\overline{\alpha}_t}X_{0}}(\sqrt{\overline{\alpha}_t}x_0)} \notag\\
	&= -\frac{\int_{x_0}\frac{\sqrt{\overline{\alpha}_t}x_0-x}{1-\overline{\alpha}_t}\exp\Big(-\frac{\|x-\sqrt{\overline{\alpha}_t}x_0\|^2}{2(1-\overline{\alpha}_t)}\Big)\mathrm{d}P_{\sqrt{\overline{\alpha}_t}X_{0}}(\sqrt{\overline{\alpha}_t}x_0)}{\int_{x_0}\exp\Big(-\frac{\|x-\sqrt{\overline{\alpha}_t}x_0\|^2}{2(1-\overline{\alpha}_t)}\Big)\mathrm{d}P_{\sqrt{\overline{\alpha}_t}X_{0}}(\sqrt{\overline{\alpha}_t}x_0)}\notag \\
	&= \mathbb{E}_{X_0 \mymid X_t =x}\Big[\frac{\sqrt{\overline{\alpha}_t}X_0-x}{1-\overline{\alpha}_t}\Big].
	\label{eqn:score-violin}
\end{align}
The second moment of score function thus can be written as
\begin{align*}
	\mathbb{E}\|s_t(X_t)\|^2 &= \mathbb{E}_{X_t}\Big\|\mathbb{E}_{X_0 \mymid X_t}\Big[\frac{\sqrt{\overline{\alpha}_t}X_0-X_t}{1-\overline{\alpha}_t}\Big]\Big\|_2^2\\
	&\le \mathbb{E}_{X_t}\Big[\mathbb{E}_{X_0 \mymid X_t}\Big\|\frac{\sqrt{\overline{\alpha}_t}X_0-X_t}{1-\overline{\alpha}_t}\Big\|_2^2\Big]\\
	&= \mathbb{E}_{X_t}\Big[\frac{1}{(1-\overline{\alpha}_t)^2}\mathbb{E}_{X_0 \mymid X_t}\|\sqrt{\overline{\alpha}_t}X_0-X_t\|_2^2\Big]\\
	&= \frac{d}{1-\overline{\alpha}_t},
\end{align*}
where the last line makes use of the expression~\eqref{eqn:Xt-X0}. 

\subsection{Proof of Lemma~\ref{lemma:discrete}}
\label{sec:pf-lem-discrete}

Throughout this proof, we adopt the truncation strategy onto the typical event $\mathcal{E}_{t}$ (defined in expression~\eqref{eqn:eset}). The targeted expectation is then calculated by considering the typical event and its complement separately.  

\paragraph{On the typical event $\mathcal{E}_t$.}
Let us first consider the case when $(x_t,x_{t-1}) \in \mathcal{E}_t$.
We claim that 
\begin{align}\label{eq:lemma2-con} 
	\|s_{\overline{\alpha}}(g_t(x_t,\overline{\alpha}))\|_2^2 \le c_5\frac{d\log T}{1 - \overline{\alpha}_t}\quad \text{and} \quad \Big\|g_t(x_t,\overline{\alpha}) - \sqrt{\frac{\overline{\alpha}}{\overline{\alpha}_t}}x_t\Big\|_2 \le c_6\sqrt{d(1-\alpha_t)\log T}
\end{align}
hold for all $\overline{\alpha}_t \le \overline{\alpha} \le \overline{\alpha}_{t-1}$.
This claim essentially means that every $(x_t,x_{t-1}) \in \mathcal{E}_t$ induces a trajectory on which all the points share similar properties as the definition of $\mathcal{E}_{t}$.
In the following proof, we shall use $\widetilde{\alpha}$ as the variable of integration to differentiate from $\overline{\alpha}$, which serves as an argument.

Before proceeding, we isolate some properties obtained with the help of this claim.
In particular, if relation~\eqref{eq:lemma2-con} holds, then dynamic~\eqref{eq:tranformed ODE} implies that
\begin{align}
g_t(x_t,\overline{\alpha}) = & \sqrt{\overline{\alpha}}\Big(\frac{x_t}{\sqrt{\overline{\alpha}_t}} + \frac{1}{2}\int_{\overline{\alpha}_t}^{\overline{\alpha}}\sqrt{\frac{1}{\widetilde{\alpha}^3}}\big(s_{\widetilde{\alpha}}(g_t(x_t,\widetilde{\alpha})) - s_t(x_{t})\big)\mathrm{d}\widetilde{\alpha}\Big)\notag\\
=& \sqrt{\overline{\alpha}}\Big(\frac{x_t}{\sqrt{\overline{\alpha}_t}} + \frac{1}{2}\int_{\overline{\alpha}_t}^{\overline{\alpha}}\sqrt{\frac{1}{\widetilde{\alpha}^3}}\mathrm{d}\widetilde{\alpha}\cdot O\big(\sup_{\overline{\alpha}_t < \widetilde{\alpha} < \overline{\alpha}}\|s_{\widetilde{\alpha}}(g_t(x_t,\widetilde{\alpha})) - s_t(x_{t})\|_2\big)\Big)\notag\\
\le& \sqrt{\frac{\overline{\alpha}}{\overline{\alpha}_t}}x_t + O\bigg(\sqrt{\overline{\alpha}_{t-1}}\Big(\frac{1}{\sqrt{\overline{\alpha}_t}}-\frac{1}{\overline{\alpha}_{t-1}}\Big)\sup_{\overline{\alpha}_t < \widetilde{\alpha} < \overline{\alpha}}\|s_{\widetilde{\alpha}}(g_t(x_t,\widetilde{\alpha}))\|_2\bigg)\notag\\
=& \sqrt{\frac{\overline{\alpha}}{\overline{\alpha}_t}}x_t + O\bigg((1 - \alpha_t )\sup_{\overline{\alpha}_t < \widetilde{\alpha} < \overline{\alpha}}\|s_{\widetilde{\alpha}}(g_t(x_t,\widetilde{\alpha}))\|_2\bigg)\notag\\
=& \sqrt{\frac{\overline{\alpha}}{\overline{\alpha}_t}}x_t + O\bigg(\sqrt{\frac{d(1-\alpha_t)^2\log T}{1 - \overline{\alpha}_t}}\bigg),\label{eq:diff-x}
\end{align}
where the last line holds using the bound~\eqref{eq:lemma2-con}. 
In addition, given the claim~\eqref{eq:lemma2-con}, according to (161c) in~\citet[Appendix C.1]{li2023towards}, the following inequality holds:
\begin{align}\label{eq:s-diff-2norm}
	\|s_{\overline{\alpha}}(g_t(x_t,\overline{\alpha})) - s_t(x_t)\|_2 \lesssim (1-\alpha_t)\bigg(\frac{d\log T}{1 - \overline{\alpha}_t}\bigg)^{3/2}.
\end{align}

\paragraph{Proof of relation~\eqref{eq:lemma2-con}.} 
We establish the relation~\eqref{eq:lemma2-con} by contradiction.
If the condition does not hold along the trajectory, let us define
\begin{align*}
	\widehat{\alpha} := \min\Big\{\overline{\alpha} : \|s_{\overline{\alpha}}(g_t(x_t,\overline{\alpha}))\|_2^2 > \frac{c_5 d\log T}{1 - \overline{\alpha}_t} \enspace \text{or} \enspace \|g_t(x_t,\overline{\alpha}) - \sqrt{\overline{\alpha}/\overline{\alpha}_t}x_t\|_2 > c_6\sqrt{d(1-\alpha_t)\log T}\Big\}.
\end{align*}
The contradiction appears if we show both scenarios in the definition of $\widehat{\alpha}$ cannot happen.
By virtue of this definition, it satisfies that for $\overline{\alpha}_t \le \widehat{\alpha} < \overline{\alpha}$, inequalities~\eqref{eq:diff-x} and \eqref{eq:s-diff-2norm} still hold true.

\begin{itemize}
\item If for the defined $\widehat{\alpha}$, we have $\|g_t(x_t,\overline{\alpha}) - \sqrt{\overline{\alpha}/\overline{\alpha}_t}x_t\|_2 > c_6\sqrt{d(1-\alpha_t)\log T}$,
Then, by calculations in expression~\eqref{eq:diff-x}, $g_t(x_t,\widehat{\alpha})$ can be written as 
\begin{align*}
	g_t(x_t,\widehat{\alpha}) = & \sqrt{\widehat{\alpha}}\Big(\frac{x_t}{\sqrt{\overline{\alpha}_t}} + \frac{1}{2}\int_{\overline{\alpha}_t}^{\widehat{\alpha}}\sqrt{\frac{1}{\widetilde{\alpha}^3}}\big(s_{\widetilde{\alpha}}(g_t(x_t,\widetilde{\alpha})) - s_t(x_{t})\big)\mathrm{d}\widetilde{\alpha}\Big)\\
	=& \sqrt{\frac{\widehat{\alpha}}{\overline{\alpha}_t}}x_t + O\bigg(\sqrt{\frac{d(1-\alpha_t)^2\log T}{1 - \overline{\alpha}_t}}\bigg)\\
	\le& \sqrt{\frac{\widehat{\alpha}}{\overline{\alpha}_t}}x_t + O(\sqrt{d(1-\alpha_t)\log T}).
\end{align*}
which is contradicted with the assumption $\|g_t(x_t,\overline{\alpha}) - \sqrt{\overline{\alpha}/\overline{\alpha}_t}x_t\|_2 > c_5\sqrt{d(1-\alpha_t)\log T}$.

\item Otherwise, consider the case that $\|s_{\widehat{\alpha}}(g_t(x_t,\widehat{\alpha}))\|_2^2 > \frac{c_5 d\log T}{1 - \overline{\alpha}_t}$. 
For $\overline{\alpha}_t \le \widehat{\alpha} < \overline{\alpha}$, by inequality~\eqref{eq:s-diff-2norm}, we directly obtain
\begin{align*}
	\|s_{\overline{\alpha}}(g_t(x_t,\overline{\alpha}))\|_2 \le O\bigg((1-\alpha_t)\bigg(\frac{d\log T}{1 - \overline{\alpha}_t}\bigg)^{3/2}\bigg) + O\bigg(\frac{d\log T}{1 - \overline{\alpha}_t}\bigg)^{1/2} = O\bigg(\frac{d\log T}{1 - \overline{\alpha}_t}\bigg)^{1/2},
\end{align*}
where we use the fact that
\begin{align}
	\|s_t(x_t)\|^2 \lesssim \frac{d\log T}{1-\overline{\alpha}_t},\label{eq:s-E} 
\end{align}
whose proof can be found as in (128b) of~\citet[Appendix B.1.1]{li2023towards}.
We can then make use of the continuity of $s_\alpha(x)$ and trajectory to obtain $\|s_{\widehat{\alpha}}(g_t(x_t,\widehat{\alpha}))\|_2 \lesssim (\frac{d\log T}{1 - \overline{\alpha}_t})^{1/2}$.
This result is also contradicted with the definition of $\widehat{\alpha}$.
%Here, we make use of the continuity of 
%\begin{align*}
%s_{\overline{\alpha}}(x) &= -\frac{1}{\sqrt{1-\overline{\alpha}}}\int p_{X_0\mymid g_t(x_t,\overline{\alpha})}(x_0 \mymid x)(x - \sqrt{\overline{\alpha}}x_0)\mathrm{d} x_0 \\
%&= -\frac{1}{\sqrt{1-\overline{\alpha}_t}}\int p_{X_0\mymid X_t}(x_0 \mymid x)(x - \sqrt{\overline{\alpha}_t}x_0)\mathrm{d} x_0
%\end{align*}
\end{itemize}
Putting everything together, we conclude that $\widehat{\alpha} \in [\overline{\alpha}_{t}, \overline{\alpha}_{t-1}]$ does not exist, which thus validates the claim~\eqref{eq:lemma2-con}.

\paragraph{On the complement of the typical event $\mathcal{E}_t^{c}$.} 
Let us now turn to the case when $(x_t,x_{t-1}) \in \mathcal{E}_t^c$.
% First, the definition of the score function $s_{\overline{\alpha}}(x)$ directly leads to
% \begin{align}
% 	\|s_t(x)\|_2 =& \bigg\|\frac{\nabla_x p_{X_t}(x)}{p_{X_t}(x)}\bigg\|_2\notag\\
% 	=& \Bigg\|\frac{\nabla_x \int_{x_0} p_{X_0}(x_0) \exp\Big(-\frac{\|x - \sqrt{\overline{\alpha}_t}x_0\|_2^2}{2(1-\overline{\alpha}_t)}\Big)\mathrm{d}x_{0}}{\int_{x_0} p_{X_0}(x_0) \exp\Big(-\frac{\|x - \sqrt{\overline{\alpha}_t}x_0\|_2^2}{2(1-\overline{\alpha}_t)}\Big)\mathrm{d}x_{0}}\Bigg\|_2\notag\\
% 	\le& \sup_{x_0:\|x_0\|_2 \le R} \frac{\|x-\sqrt{\overline{\alpha}_t}x_0\|_2}{1-\overline{\alpha}_t }\notag\\
% 	\le&\sup_{x_0:\|x_0\|_2 \le R} \frac{\|x\|_2+\|\sqrt{\overline{\alpha}_t}x_0\|_2}{1-\overline{\alpha}_t},\label{eq:st-out-e}
% \end{align}
% where $R =T^{c_R}$ defined in equation $\eqref{eqn:boundness}$. 
% \yuting{can you directly use relation~\eqref{eqn:score-violin}; recall here the definition of $R$.}
% \zh{we can use relation~\eqref{eqn:score-violin} with some duduction, the bound is the same. I rewrite the following proof in this way.}
% In Lemma~\ref{lemma:discrete}, we are actually interest in the expectation of the truncation error.
Using the upper bound in Lemma~\ref{lemma:score moment}, we integrate over the tail event of $X_t$ and $X_{t-1}$ as inequality \eqref{eq:P-Xt-X-t-1-not-E} to obtain
\begin{align}
	\mathbb{E}_{X_t,X_{t-1}}\Big[\|s_t(X_t)\|_2^2 \bm 1\big((X_t,X_{t-1}) \in \mathcal{E}_t^c\big) \Big]
	\lesssim &\int_{\mathcal{E}_t^c}\|s_t(X_t)\|_2^2p_{X_{t-1},X_{t}}(x_{t-1}, x_{t})\mathrm{d}x_{t-1}\mathrm{d}x_t\notag\\
	\lesssim &\int_{\mathcal{E}_t^c}\|s_t(X_t)\|_2^2p_{X_{t-1}\mymid X_{t}}(x_{t-1}\mymid x_{t})p_{X_{t}}(x_{t})\mathrm{d}x_{t-1}\mathrm{d}x_t\notag\\
	\lesssim &\frac{d}{1-\overline{\alpha}_t}\int_{x_{t-1} : (x_t,x_{t-1}) \in \mathcal{E}_t^c}p_{X_{t-1}\mymid X_{t}}(x_{t-1}\mymid x_{t})\mathrm{d}x_{t-1}.
	% \notag\\
	% \lesssim &\exp(-c_4 d \log T),
	% \label{eq:E-out-E}
\end{align}
% where the last line use the fact that $X_{t-1}|X_{t}$ follows a Gaussian distribution, and can be bounded by truncation condition.
% Specifically, 
It has been shown in~\citet[Step 3, Appendix C.1]{li2023towards} that
% \begin{align*}
% 	\int_{x_{t-1} : (x_t,x_{t-1}) \in \mathcal{E}_t}p_{X_{t-1}\mymid X_{t}}(x_{t-1}\mymid x_{t}) d x_{t-1}  \lesssim (1 - \exp(-c_4 d \log T))\Big(1 + O\Big(\frac{d^2 (1-\alpha_t)\log^2 T}{\alpha_t(1-\overline{\alpha}_{t-1})}\Big)\Big),
% \end{align*}
% which directly gives us 
\begin{align*}
	\int_{x_{t-1} : (x_t,x_{t-1}) \in \mathcal{E}_t^c}p_{X_{t-1}\mymid X_{t}}(x_{t-1}\mymid x_{t}) d x_{t-1}  \lesssim \exp(-c_4 d \log T).
\end{align*}
By virtue of this relation, we can conclude that 
\begin{align}
	\mathbb{E}_{X_t,X_{t-1}}\big[\|s_t(X_t)\|_2^2 \bm 1\big((X_t,X_{t-1}) \in \mathcal{E}_t^c\big) \big]
	\lesssim & \exp(-c_4 d \log T).
	\label{eq:E-out-E}
\end{align}
% \yuting{TBA}\zh{done}
% Here, we combine the tail probability \eqref{eq:P-Xt-X-t-1-not-E} with some direct calculations. 
% Similarly, for $s_{\overline{\alpha}}(g_t(x_t,\overline{\alpha}))$, one can obtain 
% \begin{align*}
% 	\|s_{\overline{\alpha}}(g_t(x_t,\overline{\alpha}))\|_2 \leq \sup_{x_0:\|x_0\|_2 \le R} \frac{\|g_t(x_t,\overline{\alpha})\|_2+\|\sqrt{\overline{\alpha}}x_0\|_2}{1-\overline{\alpha}}.
% \end{align*}

Here, similar to the proof of Lemma~\ref{lemma:score moment}, it holds that 
% \begin{align*} 
% \mathbb{E}_{X_t}\|s_{\overline{\alpha}}(g_t(X_t,\overline{\alpha}))\|_2^2 
% &\le 2\mathbb{E}_{X_t}\Big[\frac{1}{(1-\overline{\alpha})^2}\mathbb{E}_{X_0 \mymid X_t}\big[\|\sqrt{\overline{\alpha}_t}X_0-X_t\|_2^2\\ 
% &\qquad\qquad\qquad\qquad\qquad\qquad+ \|(\sqrt{\overline{\alpha}} - \sqrt{\overline{\alpha}_t})X_0 + X_t - g_t(X_t,\overline{\alpha})\|_2^2\big]\Big]\\
% &\lesssim \frac{d}{1-\overline{\alpha}_t} + \mathbb{E}_{X_0}\|(\sqrt{\overline{\alpha}_{t-1}} - \sqrt{\overline{\alpha}_t})X_0\|_2^2 + \mathbb{E}_{X_t}\Big[\|X_t\|_2^2 + \|g_t(X_t,\overline{\alpha})\|_2^2\Big]\\
% &\lesssim \frac{d}{1-\overline{\alpha}_t} + \mathbb{E}_{X_0}\|(\sqrt{\overline{\alpha}_{t-1}} - \sqrt{\overline{\alpha}_t})X_0\|_2^2 + \mathbb{E}_{X_t}\Big[\|X_t\|_2^2 + \|g_t(X_t,\overline{\alpha})\|_2^2\Big]
% &\lesssim \frac{d}{1-\overline{\alpha}_t} + \frac{T^{2c_R}}{(1-\overline{\alpha}_t)^2},
% \end{align*}
\begin{align*}
	\mathbb{E}_{X_t}\|s_{\overline{\alpha}}(g_t(X_t,\overline{\alpha}))\|_2^2 
	= \mathbb{E}_{X(\overline{\alpha})}\|s_{\overline{\alpha}}(X(\overline{\alpha}))\|_2^2	
	\le \frac{d}{1-\overline{\alpha}},
\end{align*}
where use the fact that $g_t(X_t,\overline{\alpha}) \overset{\mathsf{d}}{=} X(\overline{\alpha})$.
% \yuting{more details, where does $s_t(X_t)$ come from? And $\eqref{eq:diff-x}$ is on event $\mathcal{E}_{t}$?}\zh{done. rewrite it to be easier to understand}
As a result, this inequality enables us to bound the expectation of the truncation error in a similar way as in inequality~\eqref{eq:E-out-E}:
\begin{align}
	\mathbb{E}_{X_t,X_{t-1}}\Big[\|s_{\overline{\alpha}}(g_t(X_t,\overline{\alpha}))\|_2^2 \bm 1\big((X_t,X_{t-1}) \in \mathcal{E}_t^c\big)\Big]
	%\lesssim &\int_{\mathcal{E}^c}T^2(\|g_t(x_t,\overline{\alpha})\|_2^2 + T^{2c_R})p_{X_{t-1},X_{t}}(x_{t-1}, x_{t})\mathrm{d}x_{t-1}\mathrm{d}x_t \notag \\
	%\lesssim &\exp(-c_3 d \log T) + \int_{\mathcal{E}^c}T^2\|g_t(x_t,\overline{\alpha})\|_2^2 p_{X_{t-1},X_{t}}(x_{t-1}, x_{t})\mathrm{d}x_{t-1}\mathrm{d}x_t\notag\\
	\lesssim &\int_{\mathcal{E}_t^c}\|s_{\overline{\alpha}}(g_t(X_t,\overline{\alpha}))\|_2^2 p_{X_{t-1}\mymid X_{t}}(x_{t-1}\mymid x_{t})p_{X_{t}}(x_{t})\mathrm{d}x_{t-1}\mathrm{d}x_t\notag\\
	\lesssim &\frac{d}{1-\overline{\alpha}_t}\int_{x_{t-1} : (x_t,x_{t-1}) \in \mathcal{E}_t^c}p_{X_{t-1}\mymid X_{t}}(x_{t-1}\mymid x_{t})\mathrm{d}x_{t-1}\notag\\
	\lesssim &\exp(-c_4 d \log T).\notag
\end{align}

\paragraph{In summary.} Combining the two cases above, we conclude that
\begin{align*}
&\mathbb{E}\Big[\Big\|\int_{\overline{\alpha}_t}^{\overline{\alpha}_{t-1}}\sqrt{\frac{\overline{\alpha}_t}{\overline{\alpha}^3}}\big(s_{\overline{\alpha}}(g_t(X_t,\overline{\alpha})) - s_t(X_{t})\big)\mathrm{d}\overline{\alpha}\Big\|_2^2\Big]\\ 
& \le \int_{\overline{\alpha}_t}^{\overline{\alpha}_{t-1}}\sqrt{\frac{\overline{\alpha}_t}{\overline{\alpha}^3}}\mathbb{E}\big[\|s_{\overline{\alpha}}(g_t(X_t,\overline{\alpha})) - s_t(X_{t})\|_2^2\big]\mathrm{d}\overline{\alpha} \\
&= \int_{\overline{\alpha}_t}^{\overline{\alpha}_{t-1}}\sqrt{\frac{\overline{\alpha}_t}{\overline{\alpha}^3}}\mathbb{E}\Big[\|s_{\overline{\alpha}}(g_t(X_t,\overline{\alpha})) - s_t(X_{t})\|_2^2\big(\bm 1\big((X_t,X_{t-1})\in\mathcal{E}_t\big) + \bm 1\big((X_t,X_{t-1})\in\mathcal{E}_t^c\big)\big)\Big] \mathrm{d}\overline{\alpha}\\
&\lesssim \frac{(1 - \alpha_t)^4d^3\log^3 T}{(1-\overline{\alpha}_{t})^3} + \exp(-c_4 d \log T)\\
&\lesssim \frac{(1 - \alpha_t)^4d^3\log^3 T}{(1-\overline{\alpha}_{t})^3},
\end{align*}
which thus validates the claimed result.

\subsection{Proof of Claim~\eqref{eq:Jacobi-all}}
\label{sec:pf-jacobi}

Towards this, let us first make the observation that
\begin{align*}
\frac{\partial \Phi_{t \to k}}{\partial x}(x) &= \frac{\partial \Phi_{t-1 \to k}}{\partial x}\big(\phi_t(x)\big)\frac{\partial \phi_t}{\partial x}(x) \\
&= \frac{\partial \Phi_{t-2 \to k}}{\partial x}\big(\Phi_{t \to t-2}(x)\big)\frac{\partial \phi_{t-1}}{\partial x}\big(\Phi_{t \to t-1}(x)\big)\frac{\partial \phi_t}{\partial x}(x) \\
&= \frac{\partial \Phi_{k^{\prime} \to k}}{\partial x}\big(\Phi_{t \to k^{\prime}}(x)\big)\prod_{i=k}^{k^{\prime}} \frac{\partial \phi_{i}}{\partial x}\big(\Phi_{t \to i}(x)\big) \\
&= \prod_{i = k+1}^t \frac{\partial \phi_{i}}{\partial x}\big(\Phi_{t \to i}(x)\big),
\end{align*}
% \yuting{k+1 or k?}\zh{done}
where we recursively apply the definition of $\Phi_{k^{\prime} \to k} = \phi_{k^{\prime}} \circ \Phi_{k^{\prime} - 1 \to k}$. 
In view of the relation above, by some direct algebra, we deduce 
\begin{align*}
\bigg\|\frac{\partial \Phi_{t}}{\partial x}(x) - \frac{\partial \Phi_{t}}{\partial x}(y)\bigg\| &= \bigg\|\prod_{i = 2}^t \frac{\partial \phi_{i}}{\partial x}\big(\Phi_{t \to i}(x)\big) - \prod_{i = 2}^t \frac{\partial \phi_{i}}{\partial x}\big(\Phi_{t \to i}(y)\big)\bigg\| \\
&=\bigg\|\sum_{k = 3}^{t} \bigg(\prod_{i = 2}^{k-1} \frac{\partial \phi_{i}}{\partial x}\big(\Phi_{t \to i}(x)\big)
\bigg(\frac{\partial \phi_{k}}{\partial x}\big(\Phi_{t \to k}(x)\big) - \frac{\partial \phi_{k}}{\partial x}\big(\Phi_{t \to k}(y)\big)\bigg)
\prod_{i = k+1}^t \frac{\partial \phi_{i}}{\partial x}\big(\Phi_{t \to i}(y)\big)\bigg)\bigg\| \\
&\le \sum_{k = 2}^{t} \bigg\|\prod_{i = 2}^{k-1} \frac{\partial \phi_{i}}{\partial x}\big(\Phi_{t \to i}(x)\big)\bigg\|
\bigg\|\frac{\partial \phi_{k}}{\partial x}\big(\Phi_{t \to k}(x)\big) - \frac{\partial \phi_{k}}{\partial x}\big(\Phi_{t \to k}(y)\big)\bigg\|
\bigg\|\prod_{i = k+1}^t \frac{\partial \phi_{i}}{\partial x}\big(\Phi_{t \to i}(y)\big)\bigg\| \\
&= \sum_{k = 2}^{t} \bigg\|\frac{\partial \Phi_{k-1}}{\partial x}\big(\Phi_{t \to k-1}(x)\big)\bigg\|
\bigg\|\frac{\partial \phi_{k}}{\partial x}\big(\Phi_{t \to k}(x)\big) - \frac{\partial \phi_{k}}{\partial x}\big(\Phi_{t \to k}(y)\big)\bigg\|
\bigg\|\frac{\partial \Phi_{t\to k}}{\partial x}(y)\bigg\| \\
&\le L_f^2\sum_{k = 2}^{t} \bigg\|\frac{\partial \phi_{k}}{\partial x}\big(\Phi_{t \to k}(x)\big) - \frac{\partial \phi_{k}}{\partial x}\big(\Phi_{t \to k}(y)\big)\bigg\|.
\end{align*}
where we denote $\prod_{i}^{i-1} (\partial{\phi_i}/\partial x) := 1$ for saimplicity, and the last invokes the Assumption~\ref{ass:Lipschitz} again. 

\subsection{Proof of Lemma~\ref{lemma:Jacobi-diff}}
\label{sec:pf-jacobi}

% Using the same strategy, we first consider the case that $(x_k,x_{k-1}) \in \mathcal{E}_k$.

To begin with, let us first provide a more succinct expression for quantity $\frac{\partial \phi_k(x)}{\partial x}$. 
Recall that $\phi_k(x) := g_{k}(x, \overline{\alpha}_{k-1}).$ 
In view of relation~\eqref{eq:tranformed ODE}, we can write 
\begin{align}
 \phi_k(x) 
 &= \sqrt{\overline{\alpha}_{k-1}}\Bigg(\frac{g_k(x,\overline{\alpha}_k)}{\sqrt{\overline{\alpha}_k}} + \int_{\overline{\alpha}_k}^{\overline{\alpha}_{k-1}} \frac{1}{2\overline{\alpha}^{\frac{3}{2}}}s_{\overline{\alpha}}(g_k(x,\overline{\alpha}))\mathrm{d}\overline{\alpha}\Bigg) \notag \\
&= \frac{1}{\sqrt{{\alpha}_{k}}}\Bigg(x + \frac{1}{2}\int_{\overline{\alpha}_k}^{\overline{\alpha}_{k-1}} \sqrt{\frac{\overline{\alpha}_k}{\overline{\alpha}^3}} s_{\overline{\alpha}}(g_k(x,\overline{\alpha}))\mathrm{d}\overline{\alpha}\Bigg).
\end{align}
By some direct calculations, we arrive at 
\begin{align}
\label{eq:derivative-phi}
	\frac{\partial \phi_k(x)}{\partial x} &= \frac{1}{\sqrt{\alpha_{k}}}\bigg(I + \frac{1}{2}\int_{\overline{\alpha}_k}^{\overline{\alpha}_{k-1}}\sqrt{\frac{\overline{\alpha}_k}{\overline{\alpha}^3}}\nabla s_{\overline{\alpha}}(g_k(x,\overline{\alpha}))\frac{\partial g_k(x,\overline{\alpha})}{\partial x}\mathrm{d}\overline{\alpha}\bigg)
\end{align}
where we write $\nabla s_{\overline{\alpha}}(g_k(x,\overline{\alpha})) := \nabla_y s_{\overline{\alpha}}(y)\big|_{y = g_k(x,\overline{\alpha})}$.
We then proceed to control each term in the above expression. To do so, let us introduce the following two lemmas whose proofs are provided in Section~\ref{sec:pf-lem-der-s} and \ref{sec:pf-lem-der-x} respectively. 

\begin{lemma}\label{lemma:derivative-s}
	For $2 \le t \le T$, $\overline{\alpha}_t \le \overline{\alpha} \le \overline{\alpha}_{t-1}$ and $(x_t,x_{t-1}) \in \mathcal{E}_t$, the derivative of the score function satisfies
	\begin{align*}
		\big\|\nabla s_{\overline{\alpha}}(g_t(x_t,\overline{\alpha})) - \nabla s_{t}(x_t)\big\|_2 \lesssim \frac{d^2(1-\alpha_t)\log^2 T}{(1 - \overline{\alpha}_t)^2}.
	\end{align*}
\end{lemma}
\begin{lemma}\label{lemma:derivative-x}
	For $2 \le t \le T$ and $(x_t,x_{t-1}) \in \mathcal{E}_t$, the stability of the backward ODE~\eqref{eq:tranformed ODE} starting at $x_t$ can be bounded as follows:
	\begin{align*}
		\bigg\| \frac{\partial g_t(x_t,\overline{\alpha})}{\partial x} - I \bigg\|\lesssim \frac{d(1-\alpha_t)\log T}{1-\overline{\alpha}_t}.
	\end{align*}
\end{lemma}

Plugging in the bounds from Lemma~\ref{lemma:derivative-s} and Lemma~\ref{lemma:derivative-x} to equation~\eqref{eq:derivative-phi}, we obtain 
\begin{align}
\frac{\partial \phi_k(x)}{\partial x} &= \frac{1}{\sqrt{\alpha_{k}}}\bigg(I + \frac{1}{2}\int_{\overline{\alpha}_k}^{\overline{\alpha}_{k-1}}\sqrt{\frac{\overline{\alpha}_k}{\overline{\alpha}^3}}\Big(\frac{\partial s_{k}(x)}{\partial x}+O\Big(\frac{d^2(1-\alpha_k)\log^2 T}{(1 - \overline{\alpha}_k)^2}\Big)\Big)\Big(I_d + O\Big(\frac{d(1-\alpha_k)\log T}{1-\overline{\alpha}_t}\Big)\Big)\mathrm{d}\overline{\alpha}\bigg)\notag\\
&= \frac{1}{\sqrt{\alpha_{k}}}\bigg(I + \frac{1}{2}\int_{\overline{\alpha}_k}^{\overline{\alpha}_{k-1}}\sqrt{\frac{\overline{\alpha}_k}{\overline{\alpha}^3}}\frac{\partial s_{k}(x)}{\partial x}\mathrm{d}\overline{\alpha}\bigg) + O\bigg(\frac{d^2(1-\alpha_k)^2\log^2 T}{(1 - \overline{\alpha}_k)^2}\bigg)\notag\\
&= \frac{1}{\sqrt{\alpha_{k}}}\bigg(I-\frac{1-\sqrt{\alpha_{k}}}{1-\overline{\alpha}_{k}}J_k(x)\bigg) + O\bigg(\frac{d^2(1-\alpha_k)^2\log^2 T}{(1 - \overline{\alpha}_k)^2}\bigg),\label{eq:Jt-st}
\end{align}
where we denote 
\begin{align}
J_{k}(x) & :=I_{d}+\frac{1}{1-\overline{\alpha}_{k}}\bigg\{\mathbb{E}\big[X_{k}-\sqrt{\overline{\alpha}_{k}}X_{0}\mid X_{k}=x\big]\Big(\mathbb{E}\big[X_{k}-\sqrt{\overline{\alpha}_{k}}X_{0}\mid X_{k}=x\big]\Big)^{\top}\notag\\
 & \qquad\qquad\qquad-\mathbb{E}\Big[\big(X_{k}-\sqrt{\overline{\alpha}_{k}}X_{0}\big)\big(X_{k}-\sqrt{\overline{\alpha}_{k}}X_{0}\big)^{\top}\mid X_{k}=x\Big]\bigg\}.
	\label{eq:Jt-expression}
\end{align}
The details for deriving expression~\eqref{eq:Jt-st} are included in Section~\ref{sec:jt-st}.

%Then for
%\begin{align}
%	\theta_t(x, y) \coloneqq \max\bigg\{\sup_{0 \le \gamma \le 1} \frac{-\log {p_{X_t}(\gamma x + (1-\gamma)y)}}{d\log T} , c_6 \bigg\},
%\end{align}
%it can be checked that
%\begin{align}
%\|J_t(x) - J_t(y)\| \lesssim L_s\sqrt{\theta_t(x, y) d\log T}\|x - y\|_2.
%\end{align}

In order to prove Lemma~\ref{lemma:Jacobi-diff} and cope with the difference $\frac{\partial \phi_{k}}{\partial x}\big(\Phi_{t-1 \to k}(X_{t-1}(\gamma))\big) - \frac{\partial \phi_{k}}{\partial x}\big(\Phi_{t \to k}(X_t)\big)$,
inequality~\eqref{eq:Jt-st} suggests to study the Lipschitz property of function $J_k.$
For this purpose, we introduce our final auxiliary result, whose proof is provided in Section~\ref{sec:pf-lem-J}. 
\begin{lemma}\label{lemma:J-Lip}
	For $2 \le t \le T$ and $(x_t,x_{t-1}) \in \mathcal{E}_t$, $J_t(x)$ is locally Lipschitz continuous with respect to $x$:
	\begin{subequations}
	\begin{align}
	\label{eqn:J-lip-1}
		\|J_t(x_{t-1}) - J_t(\phi(x_t))\| \lesssim \frac{1}{\sqrt{1-\overline{\alpha}_t}}d^{3/2}\log^{3/2} T \|x_{t-1} - \phi(x_t)\|_2.
	\end{align}
	In addition, for $1\le k \le t-1$, the Lipschitz constant along the backward trajectory satisfies 
	\begin{align}
	\label{eqn:J-lip-2}
		\Big\|J_k\big(\Phi_{t-1 \to k}(x_{t-1}(\gamma))\big) - J_k\big(\Phi_{t \to k}(x_t)\big)\Big\| \lesssim \frac{1}{\sqrt{1-\overline{\alpha}_t}}d^{3/2}\log^{3/2} T\Big\|\Phi_{t-1 \to k}(x_{t-1}(\gamma)) - \Phi_{t \to k}(x_t)\Big\|_2.
	\end{align}
	\end{subequations}
\end{lemma}
% \yuting{is it sufficient to have $k \le j \le t$ and $(x_j,x_{j-1}) \in \mathcal{E}_j$?}\zh{done}

To proceed, let us again decompose the quantity of interest as  
\begin{align}
&\mathbb{E}\bigg[\bigg\|\frac{\partial \phi_{k}}{\partial x}\big(\Phi_{t-1 \to k}(X_{t-1}(\gamma))\big) - \frac{\partial \phi_{k}}{\partial x}\big(\Phi_{t \to k}(X_t)\big)\bigg\|^2\bigg]\notag \\
&=
\mathbb{E}\bigg[\bigg\|\frac{\partial \phi_{k}}{\partial x}\big(\Phi_{t-1 \to k}(X_{t-1}(\gamma))\big) - \frac{\partial \phi_{k}}{\partial x}\big(\Phi_{t \to k}(X_t)\big)\bigg\|^2\bm 1\big((X_t,X_{t-1})\in\mathcal{E}_t\big)\bigg]
\notag \\
& \qquad + \mathbb{E}\bigg[\bigg\|\frac{\partial \phi_{k}}{\partial x}\big(\Phi_{t-1 \to k}(X_{t-1}(\gamma))\big) - \frac{\partial \phi_{k}}{\partial x}\big(\Phi_{t \to k}(X_t)\big)\bigg\|^2\bm 1\big((X_t,X_{t-1})\in\mathcal{E}_t^c\big)\bigg].
\end{align}
We shall control each term respectively. 

\paragraph{The first term.}
Taking Lemma~\ref{lemma:J-Lip} collectively with expression~\eqref{eq:Jt-st}, we obtain 
\begin{align}
&\mathbb{E}\bigg[\bigg\|\frac{\partial \phi_{k}}{\partial x}\big(\Phi_{t-1 \to k}(X_{t-1}(\gamma))\big) - \frac{\partial \phi_{k}}{\partial x}\big(\Phi_{t \to k}(X_t)\big)\bigg\|^2\bm 1\big((X_t,X_{t-1})\in\mathcal{E}_t\big)\bigg]\notag \\
\stackrel{(\text{i})}{\lesssim} & \frac{(1-\alpha_{k})^2}{(1-\overline{\alpha}_{k})^2} \mathbb{E}\Big[\Big\|J_k\big(\Phi_{t-1 \to k}(X_{t-1}(\gamma))\big) - J_k\big(\Phi_{t \to k}(X_t)\big)\Big\|^2\bm 1\big((X_t,X_{t-1})\in\mathcal{E}_t\big)\Big] + \frac{d^4(1-\alpha_k)^4\log^4 T}{(1 - \overline{\alpha}_k)^4}\notag \\
\stackrel{(\text{ii})}{\lesssim} & \frac{(1-\alpha_{k})^2d^{3}\log^{3} T}{(1-\overline{\alpha}_{k})^2(1-\overline{\alpha}_t)} 
\mathbb{E}\Big[\Big\|\Phi_{t-1 \to k}(X_{t-1}(\gamma))- \Phi_{t \to k}(X_t)\Big\|_2^2\bm 1\big((X_t,X_{t-1})\in\mathcal{E}_t\big)\Big] + \frac{d^4(1-\alpha_k)^4\log^4 T}{(1 - \overline{\alpha}_k)^4}\notag \\
\stackrel{(\text{iii})}{\lesssim} & 
\frac{(1-\alpha_{k})^2d^{3}\log^{3} T}{(1-\overline{\alpha}_{k})^2(1-\overline{\alpha}_t)} 
\mathbb{E}\Big[
L_f^2\big\|X_{t-1}(\gamma) - \phi(X_t)\big\|_2^2 \bm 1\big((X_t,X_{t-1})\in\mathcal{E}_t\big)\Big] + \frac{d^4(1-\alpha_k)^4\log^4 T}{(1 - \overline{\alpha}_k)^4}.
\end{align}
Note that, to ensure inequalities $(\text{i})$ and $(\text{ii})$, one invokes Lemma~\ref{lemma:J-Lip} which requires $(x_k,x_{k-1}) \in \mathcal{E}_k$.  We shall verify this relation momentarily. 
In $(\text{ii})$ we invoke the Lipschitz continuity of $\Phi_{t-1 \to k}$ and $\Phi_{t \to k}$ and the property that $\Phi_{t \to k}(X_t) \stackrel{d}{=} X_{t-1}$.  
To further control the right hand side above, recall that we have established the inequality \eqref{eq:second-term-1} when $(x_t,x_{t-1})$ in $\mathcal{E}_t$. As a result, we conclude that 
\begin{align}
\notag &\mathbb{E}\bigg[\bigg\|\frac{\partial \phi_{k}}{\partial x}\big(\Phi_{t-1 \to k}(X_{t-1}(\gamma))\big) - \frac{\partial \phi_{k}}{\partial x}\big(\Phi_{t \to k}(X_t)\big)\bigg\|^2\bm 1\big((X_t,X_{t-1})\in\mathcal{E}_t\big)\bigg]
\\
& \lesssim \frac{(1-\alpha_{k})^2(1-\alpha_t)^2L_f^2d^4\log^3 T}{(1-\overline{\alpha}_{k})^2(1 - \overline{\alpha}_t)^{2}} + \frac{(1-\alpha_k)^4d^4\log^4 T}{(1 - \overline{\alpha}_k)^4}.\label{eq:phi-in-e}
\end{align}

% \yuting{to be checked; does this argument gives you exponential dependence on $L_{f}$??} 
% \zh{will }
It is therefore only left for us to show that $(x_k,x_{k-1})$ in $\mathcal{E}_k$, which holds true owing to the Lipschitz property of $\Phi_{t-1 \to k}(x)$ and $\Phi_{t \to k}(x)$. 
Specifically, for every $(x_t,x_{t-1})$ in $\mathcal{E}_t$, by definition, it holds for large enough constant $c_4$ that 
\begin{align*}
	\|x_{t-1} - x_t/\sqrt{\alpha_t}\|_2 \leq c_4 \sqrt{d(1 - \alpha_t)\log T}.
\end{align*}
The Lipschitz continuity of $\Phi_{t \to k}$ also implies that $-\log p_{X_k}(x_k) \leq c_3 d\log T$ as $X_{k}\stackrel{\text{d}}{=} \Phi_{t \to k}(X_t).$
% Thus, the density condition in $\mathcal{E}_k$ can be checked directly by change of variables.
As a result, if we define
\begin{align*}
	\mathcal{E}_k^{\prime} \defn \bigg\{(x_k, x_{k-1}) \in \mathbb{R}^d \times \mathbb{R}^d ~\Big| -\log p_{X_k}(x_k) \leq c_3 d\log T, ~\|x_{k-1} - x_k/\sqrt{\alpha_k}\|_2 \leq c_4 L_f \sqrt{d(1 - \alpha_t)\log T} \bigg\},
\end{align*}
then one can check that $(\Phi_{t-1 \to k}(x_{t-1}),\Phi_{t \to k}(x_t)) \in \mathcal{E}_k^{\prime}$.
Notice that $\mathcal{E}_k$ and $\mathcal{E}_k^{\prime}$ share the same form for every $2\le k < t\le T$,
only with a different constant in the second condition, we conclude that Lemma~\ref{lemma:derivative-s}, \ref{lemma:derivative-x} and \ref{lemma:J-Lip} still hold true with slight different constants. 
Therefore, we have validated the relation~\eqref{eq:phi-in-e}.
% Therefore, when truncating on $\mathcal{E}_t$ in inequality~\eqref{eq:phi-in-e}, we can apply the former lemmas for $\phi_k(x)$ and $J_k(x)$ with truncation on $\mathcal{E}'_k$ with only a constant level change on the upper bound.

\paragraph{The second term.} When $(x_t,x_{t-1})\in\mathcal{E}^{c}$ holds true, it is sufficient to consider a crude upper bound for 
\begin{align*}
	\bigg\|\frac{\partial \phi_{k}}{\partial x}\big(\Phi_{t-1 \to k}(x_{t-1}(\gamma))\big) - \frac{\partial \phi_{k}}{\partial x}\big(\Phi_{t \to k}(x_t)\big)\bigg\|^2\bm 1\big((x_t,x_{t-1})\in\mathcal{E}_t^c\big).
\end{align*}
Owing to the Lipschitz condition in Assumption~\ref{ass:Lipschitz}, we know that
$\frac{\partial}{\partial x}\Phi_{t\to k}(x) \le L_f$.
Simply choosing $k =t-1$ gives us $\frac{\partial}{\partial x}\phi_t(x) \le L_f$, which in turn leads to
\begin{align}
	&\mathbb{E}\bigg[\bigg\|\frac{\partial \phi_{k}}{\partial x}\big(\Phi_{t-1 \to k}(X_{t-1}(\gamma))\big) - \frac{\partial \phi_{k}}{\partial x}\big(\Phi_{t \to k}(X_t)\big)\bigg\|^2\bm 1\big((X_t,X_{t-1})\in\mathcal{E}_t^c\big)\bigg]\notag\\
	\le& 4L_f^2 \mathbb{P}\big((X_{t},X_{t-1})\in\mathcal{E}_t^c\big)\notag\\
	\lesssim& L_f^2\exp(-c_4d\log T).\label{eq:phi-out-e}
\end{align}

Putting relations \eqref{eq:phi-in-e} and \eqref{eq:phi-out-e} together verifies the target result in Lemma~\ref{lemma:Jacobi-diff}.

\subsection{Proof of Claim~\eqref{eq:Jt-st}}
\label{sec:jt-st}
% The expression in~\eqref{eq:Jt-st} can be verified by the following calculation:
To establish this relation, we first find it useful to write the score function as 
\begin{align}
	s_t(x) &= \mathbb{E}\bigg[-\frac{1}{\sqrt{1-\overline{\alpha}_t}}Z~\Big|~\sqrt{\overline{\alpha}_t}X_{0}+\sqrt{1-\overline{\alpha}_t}Z = x\bigg]\notag\\
	&= -\frac{1}{1-\overline{\alpha}_t} \mathbb{E} \bigg[x - \sqrt{\overline{\alpha}_t}X_{0}~\Big|~\sqrt{\overline{\alpha}_t}x_{0}+\sqrt{1-\overline{\alpha}_t}z = x\bigg]\notag\\
	&= -\frac{1}{1-\overline{\alpha}_t} \int_{x_0}(x - \sqrt{\overline{\alpha}_t}x_{0})p_{X_0\mymid X_t}(x_0 \mymid x)\mathrm{d}x_0.
\end{align}
As a result, the partial derivative is calculated as
\begin{align}
	\frac{\partial[s_t(x)]_i}{\partial x_j} &= -\frac{1}{1-\overline{\alpha}_t}\frac{\partial}{\partial x_j}\Big[\int_{x_0}(x_i - \sqrt{\overline{\alpha}_t}x_{0,i})p_{X_0\mymid X_t}(x_0 \mymid x)\mathrm{d}x_0\Big]\notag\\
	&= -\frac{1}{1-\overline{\alpha}_t}\Big[\bm 1_{\{i=j\}} + \int_{x_0}(x_i - \sqrt{\overline{\alpha}_t}x_{0,i})\frac{\partial}{\partial x_j}p_{X_0\mymid X_t}(x_0 \mymid x)\Big]\mathrm{d}x_0\notag\\
	&= -\frac{1}{1-\overline{\alpha}_t}\Big[\bm 1_{\{i=j\}} + \int_{x_0}(x_i - \sqrt{\overline{\alpha}_t}x_{0,i})\frac{\partial}{\partial x_j}\frac{p_{X_0}(x_0)p_{X_t\mymid X_0}(x \mymid x_0)}{p_{X_t}(x)}\Big]\mathrm{d}x_0.\label{eq:J_ij}
\end{align}
By noticing the fact that 
\begin{align}\label{eq:partial-for-conditional}
	\frac{\partial}{\partial x_j} p_{X_t\mymid X_0}(x\mymid x_0) = p_{X_t\mymid X_0}(x\mymid x_0)\cdot \frac{x_j - \sqrt{\overline{\alpha}_t}x_{0,j}}{1-\overline{\alpha}_t},
\end{align}
we can thus rewrite equation~\eqref{eq:J_ij} as 
\begin{align}
	\frac{\partial[s_t(x)]_i}{\partial x_j} = & -\frac{1}{1-\overline{\alpha}_t}\Big[\bm 1_{\{i=j\}} + \frac{1}{1-\overline{\alpha}_t}\Big[\int_{x_0}(x_i - \sqrt{\overline{\alpha}_t}x_{0,i})p_{X_0\mymid X_t}(x_0 \mymid x)\mathrm{d}x_0\int_{x_0}(x_j - \sqrt{\overline{\alpha}_t}x_{0,j})p_{X_0\mymid X_t}(x_0 \mymid x)\mathrm{d}x_0\notag\\ 
	&\qquad \qquad - \int_{x_0}(x_i - \sqrt{\overline{\alpha}_t}x_{0,i})(x_j - \sqrt{\overline{\alpha}_t}x_{0,j})p_{X_0\mymid X_t}(x_0 \mymid x)\mathrm{d}x_0\Big]\Big]\label{eqn:ice-cream}\\
	&=-\frac{1}{1-\overline{\alpha}_t} [J_t(x)]_{ij}, 
\end{align}
which leads to equation~\eqref{eq:Jt-st}.

%% file: proof-auxiliary-2.tex
\subsection{Proof of Lemma~\ref{lemma:derivative-s}}
\label{sec:pf-lem-der-s}

% \yuting{the proof of this lemma is too brief}\zh{have rewritten the main part}

The proof of Claim \eqref{eq:Jt-st} provides an explicit expression of $\frac{\partial s_t(x)}{\partial x}$ via $J_t(x)$ as in expression~\eqref{eqn:ice-cream}.
% We can further write the expression of $J_t(x)$ into the integrate forms.  
% It can be computed that 
Organizing terms of expression~\eqref{eqn:ice-cream} gives us 
\begin{align*}
&\nabla s_t(x_t) +  \frac{1}{1-\overline{\alpha}_t}I_d \\ 
	=& -\frac{1}{(1-\overline{\alpha}_t)^2}\Big[\underset{\eqqcolon \, A_t }{\underbrace{\int_{x_0}(x_t - \sqrt{\overline{\alpha}_t}x_{0})p_{X_0\mymid X_t}(x_0 \mymid x_t)\mathrm{d}x_0\Big(\int_{x_0}(x_t - \sqrt{\overline{\alpha}_t}x_{0})p_{X_0\mymid X_t}(x_0 \mymid x_t)\mathrm{d}x_0\Big)^{\top}}}\\ 
	&\qquad \qquad - \underset{\eqqcolon \, B_t }{\underbrace{\int_{x_0}(x_t - \sqrt{\overline{\alpha}_t}x_{0})(x_t - \sqrt{\overline{\alpha}_t}x_{0})^{\top}p_{X_0\mymid X_t}(x_0 \mymid x_t)\mathrm{d}x_0}}\Big]
\end{align*}
and similarly, it holds that 
\begin{align*}
	&\nabla s_{\overline{\alpha}}(g_t(x_t,\overline{\alpha})) +  \frac{1}{1-\overline{\alpha}}I_d& \\
	=& -\frac{1}{(1-\overline{\alpha})^2}\Big[\underset{\eqqcolon \, A_{\overline{\alpha}} }{\underbrace{\int_{x_0}\!\!(g_t(x_t,\overline{\alpha}) - \sqrt{\overline{\alpha}}x_{0})p_{X_0\mymid X_{\overline{\alpha}}}(x_0 \mymid g_t(x_t,\overline{\alpha}))\mathrm{d}x_0\Big(\int_{x_0}\!\!(g_t(x_t,\overline{\alpha}) - \sqrt{\overline{\alpha}}x_{0})p_{X_0\mymid X_{\overline{\alpha}}}(x_0 \mymid g_t(x_t,\overline{\alpha}))\mathrm{d}x_0\Big)^{\top}}}\\ 
	&\qquad \qquad - \underset{\eqqcolon \, B_{\overline{\alpha}} }{\underbrace{\int_{x_0}(g_t(x_t,\overline{\alpha}) - \sqrt{\overline{\alpha}}x_{0})(g_t(x_t,\overline{\alpha}) - \sqrt{\overline{\alpha}}x_{0})^{\top}p_{X_0\mymid X_{\overline{\alpha}}}(x_0 \mymid g_t(x_t,\overline{\alpha}))\mathrm{d}x_0}}\Big].
\end{align*}
In view of these two decompositions, we can bound 
\begin{align}
\label{eqn:ochestra}
	\bigg\|\nabla s_{\overline{\alpha}}(g_t(x_t,\overline{\alpha})) - \nabla s_{t}(x_t)\bigg\| &\le \Big\|\frac{1}{(1-\overline{\alpha})^2}A_\alpha - \frac{1}{(1-\overline{\alpha}_t)^2}A_t\Big\| + \Big\|\frac{1}{(1-\overline{\alpha})^2}B_\alpha - \frac{1}{(1-\overline{\alpha}_t)^2}B_t\Big\|.
\end{align}
We shall proceed by controlling each term on the right respectively. 

\paragraph{Controlling the first term.} Let us start by bounding the first term. By noticing the basic algebra fact that for vectors $z_1,z_2 \in \mathbb{R}^d$,
\begin{align*}
	\|z_1 z_1^\top - z_2 z_2^\top\|_2 \leq \|z_1 - z_2\|_2\cdot \max\{\|z_1\|_2,\|z_2\|_2\},
\end{align*}
we find 
\begin{align}
	&\Big\| \frac{1}{(1-\overline{\alpha}_t)^2}A_t - \frac{1}{(1-\overline{\alpha})^2}A_\alpha\Big\|\notag\\
	&\lesssim \Big\|\frac{1}{1-\overline{\alpha}_t}\int_{x_0}(x_t - \sqrt{\overline{\alpha}_t}x_{0})p_{X_0\mymid X_t}(x_0 \mymid x_t)\mathrm{d}x_0 - \frac{1}{1-\overline{\alpha}}\int_{x_0}(g_t(x_t,\overline{\alpha}) - \sqrt{\overline{\alpha}}x_{0})p_{X_0\mymid X_{\overline{\alpha}}}(x_0 \mymid g_t(x,\overline{\alpha}))\mathrm{d}x_0\Big\|_2\cdot\notag\\
	&\qquad \qquad \qquad \qquad \qquad \qquad \qquad \qquad \qquad \qquad \qquad \qquad \qquad \qquad \Big(\max\{\|s_{\overline{\alpha}}(g_t(x,\overline{\alpha}))\|_2,\|s_t(x_t)\|_2\}\Big).\label{eq:A-diff}
\end{align}
By virtue of the bound~\eqref{eq:s-E}, we can directly derive
\begin{align}
	\max\{\|s_{\overline{\alpha}}(g_t(x,\overline{\alpha}))\|_2^2,\|s_t(x_t)\|_2^2\} \lesssim \max\Big\{\frac{d\log T}{1-\overline{\alpha}}, \frac{d\log T}{1-\overline{\alpha}_t}\Big\} = \frac{d\log T}{1-\overline{\alpha}_t}.\label{eq:diff-s3}
\end{align}
It is then sufficient to control the first term on the right hand side of inequality~\eqref{eq:A-diff}, which shall be done as follows. 
To this end, let us define a set of interest by 
\begin{align*}
	\mathcal{E}_0 \coloneqq \big\{ x: \|x_t - \sqrt{\overline{\alpha}_{t}}x\|_2 \leq c_6 \sqrt{d(1-\overline{\alpha}_t)\log T} \big\}.
\end{align*}
We first consider the the following term
\begin{align}
	&\Big\|\int_{x_0}(x_t - \sqrt{\overline{\alpha}_t}x_{0})p_{X_0\mymid X_t}(x_0 \mymid x_t)\mathrm{d}x_0 - \int_{x_0}(x_t - \sqrt{\overline{\alpha}_t}x_{0})p_{X_0\mymid X_{\overline{\alpha}}}(x_0 \mymid g_t(x,\overline{\alpha}))\mathrm{d}x_0\Big\|_2\notag\\
	&\le\int_{x_{0}\in\mathcal{E}_{0}}\big|p_{X_{0}\mymid X_{\overline{\alpha}}}(x_{0}\mymid g_t(x_t,\overline{\alpha}))-p_{X_{0}\mymid X_{t}}(x_{0}\mymid x_{t})\big|\cdot\big\| x_{t}-\sqrt{\overline{\alpha}_{t}}x_{0}\big\|_{2}\mathrm{d}x_{0}\notag\\
	& \qquad\qquad+\int_{x_{0}\in\mathcal{E}^c_{0}}\big|p_{X_{0}\mymid X_{\overline{\alpha}}}(x_{0}\mymid g_t(x_t,\overline{\alpha}))-p_{X_{0}\mymid X_{t}}(x_{0}\mymid x_{t})\big|\cdot\big\| x_{t}-\sqrt{\overline{\alpha}_{t}}x_{0}\big\|_{2}\mathrm{d}x_{0}\notag\\
	&= \int_{x_{0}\in\mathcal{E}_{0}}\bigg|\frac{p_{X_{0}\mymid X_{\overline{\alpha}}}(x_{0}\mymid g_t(x_t,\overline{\alpha}))}{p_{X_{0}\mymid X_{t}}(x_{0}\mymid x_{t})}-1\bigg|\cdot p_{X_{0}\mymid X_{t}}(x_{0}\mymid x_{t}) \cdot \big\| x_{t}-\sqrt{\overline{\alpha}_{t}}x_{0}\big\|_{2}\mathrm{d}x_{0}\notag\\
	& \qquad\qquad\int_{x_{0}\in\mathcal{E}^c_{0}}\bigg|\frac{p_{X_{0}\mymid X_{\overline{\alpha}}}(x_{0}\mymid g_t(x_t,\overline{\alpha}))}{p_{X_{0}\mymid X_{t}}(x_{0}\mymid x_{t})}-1\bigg|\cdot p_{X_{0}\mymid X_{t}}(x_{0}\mymid x_{t}) \cdot \big\| x_{t}-\sqrt{\overline{\alpha}_{t}}x_{0}\big\|_{2}\mathrm{d}x_{0}\label{eq:diff-1}.
\end{align} 
Next, we bound the right hand side above. 
Towards this, first recall that in Claim 2 in~\citet[Appendix C.1]{li2023towards}, it has been shown by direct calculations that 
\begin{align}
	\frac{p_{X_{0}\mymid X_{\overline{\alpha}}}(x_{0}\mymid g_t(x_t,\overline{\alpha}))}{p_{X_{0}\mymid X_{t}}(x_{0}\mymid x_{t})} & =1+O\bigg(\frac{d(1-\alpha_{t})\log T}{1-\overline{\alpha}_{t-1}}\bigg),\qquad
	&&\text{if }x_{0}\in\mathcal{E}_{0},\label{eq:ratio-1}\\
	\frac{p_{X_{0}\mymid X_{\overline{\alpha}}}(x_{0}\mymid g_t(x_t,\overline{\alpha}))}{p_{X_{0}\mymid X_{t}}(x_{0}\mymid x_{t})} & \leq\exp\Bigg(\frac{16c_{1}\big\| x_{t}-\sqrt{\overline{\alpha}_{t}}x_{0}\big\|_{2}^{2}\log T}{(1-\overline{\alpha}_{t})T}\Bigg),\qquad&&\text{if }x_{0}\notin\mathcal{E}_{0}.\label{eq:ratio-2}
\end{align}
Here, we remark that we replace $\sqrt{\overline{\alpha}/\overline{\alpha}_t}~x_t$ in \cite{li2023towards} by $g_t(x_t,\overline{\alpha})$.
This is valid since for $(x_t,x_{t-1}) \in \mathcal{E}_t$, inequality~\eqref{eq:diff-x} ensures 
\begin{align*}
	\Big\|\sqrt{\frac{\overline{\alpha}}{\overline{\alpha}_t}}x_t - g_t(x_t,\overline{\alpha})\Big\|_2 = O\bigg(\frac{d^{1/2}(1-\alpha_t)\log^{1/2}T}{(1-\overline{\alpha}_t)^{1/2}}\bigg).
\end{align*}
This approximation only leads to a lower order term in our final result. 
% and it can be checked that this modification will not change the order of the proven results.   
% \yuting{point out the details; the definition of $g_t$ is different in two papers.}\zh{done}

Plugging the relations~\eqref{eq:ratio-1} and \eqref{eq:ratio-2} into the right hand side of \eqref{eq:diff-1} and following the proof of (161c) in the proof in~\citet[Appendix C.1]{li2023towards}, we can obtain 
\begin{align}
	&\Big\|\int_{x_0}(x_t - \sqrt{\overline{\alpha}_t}x_{0})p_{X_0\mymid X_t}(x_0 \mymid x)\mathrm{d}x_0 - \int_{x_0}(x_t - \sqrt{\overline{\alpha}_t}x_{0})p_{X_0\mymid X_{\overline{\alpha}}}(x_0 \mymid g_t(x,\overline{\alpha}))\mathrm{d}x_0\Big\|_2\notag\\
	&\lesssim \frac{d(1-\alpha_t)\log T}{1-\overline{\alpha}_{t-1}}\cdot \mathbb{E}\left[\big\| \sqrt{\overline{\alpha}_{t}}X_{0} - x_t \big\|_{2}\,\big|\,X_{t}=x_t\right]\notag\\
	& \lesssim\frac{d^{3/2}(1-\alpha_t)\log^{3/2}T}{(1-\overline{\alpha}_t)^{1/2}},\label{eq:diff-s}
\end{align}
where we apply Lemma~\ref{lemma:x0} to deduce the last inequality. 
With the same calculations, we can similarly find
\begin{align}
	&\Big\|\int_{x_0}\!(x_t - \sqrt{\overline{\alpha}_t}x_{0})(x_t - \sqrt{\overline{\alpha}_t}x_{0})^{\top}p_{X_0\mymid X_t}(x_0 \mymid x_t)\mathrm{d}x_0 \!- \!\int_{x_0}\!(x_t - \sqrt{\overline{\alpha}_t}x_{0})(x_t - \sqrt{\overline{\alpha}_t}x_{0})^{\top}p_{X_0\mymid X_{\overline{\alpha}}}(x_0 \mymid g_t(x,\overline{\alpha}))\mathrm{d}x_0\Big\|\notag\\
	& \lesssim \frac{d(1-\alpha_t)\log T}{1-\overline{\alpha}_{t-1}}\cdot \mathbb{E}\left[\big\| \sqrt{\overline{\alpha}_{t}}X_{0} - x_t \big\|^2_{2}\,\big|\,X_{t}=x_t\right]\notag\\
	& \lesssim \frac{d^{2}(1-\alpha_t)(1-\overline{\alpha}_t)\log^{2}T}{(1-\overline{\alpha}_{t-1})} \lesssim d^2(1-\alpha_t)\log^2 T.\label{eq:diff-s2}
\end{align}

With these properties in place, we are ready to prove Lemma~\ref{lemma:derivative-x} by studying the first term in \eqref{eq:A-diff}.
First, notice that 
\begin{align*}
	&\Big\|\frac{1}{1-\overline{\alpha}_t}\int_{x_0}(x_t - \sqrt{\overline{\alpha}_t}x_{0})p_{X_0\mymid X_t}(x_0 \mymid x_t)\mathrm{d}x_0 - \frac{1}{1-\overline{\alpha}}\int_{x_0}(g_t(x_t,\overline{\alpha}) - \sqrt{\overline{\alpha}}x_{0})p_{X_0\mymid X_{\overline{\alpha}}}(x_0 \mymid g_t(x_t,\overline{\alpha}))\mathrm{d}x_0\Big\|_2\\
	&\le \frac{\sqrt{\overline{\alpha}}}{\sqrt{\overline{\alpha_{t}}}(1-\overline{\alpha})}\bigg\|\int_{x_{0}}p_{X_{0}\mymid X_{\overline{\alpha}}}(x_{0}\mymid g_t(x_t,\overline{\alpha}))(x_{t}-\sqrt{\overline{\alpha}_{t}}x_{0})\mathrm{d}x_{0}-\int_{x_{0}}p_{X_{0}\mymid X_{t}}(x_{0}\mymid x_{t})(x_{t}-\sqrt{\overline{\alpha}_{t}}x_{0})\mathrm{d}x_{0}\bigg\|_{2}\\
	& \qquad+\bigg\|\Big(\frac{\sqrt{\overline{\alpha}}}{\sqrt{\overline{\alpha}_t}(1-\overline{\alpha})}-\frac{1}{1-\overline{\alpha}_{t}}\Big)\int_{x_{0}}p_{X_{0}\mymid X_{t}}(x_{0}\mymid x_{t})(x_{t}-\sqrt{\overline{\alpha}_{t}}x_{0})\mathrm{d}x_{0}\bigg\|_{2}\\
	& \qquad\qquad +\frac{1}{1-\overline{\alpha}}\bigg\|\int_{x_0} p_{X_{0}\mymid X_{t}}(x_{0}\mymid x_{t})\bigg(g_t(x_t,\overline{\alpha}) - \sqrt{\frac{\overline{\alpha}}{\overline{\alpha}_t}}x_t\bigg)\mathrm{d}x_0\bigg\|_2\\
	&\le \frac{1}{\sqrt{\alpha_{t}}(1-\overline{\alpha}_{t-1})}\bigg\|\int_{x_{0}}p_{X_{0}\mymid X_{\overline{\alpha}}}(x_{0}\mymid g_t(x_t,\overline{\alpha}))(x_{t}-\sqrt{\overline{\alpha}_{t}}x_{0})\mathrm{d}x_{0}-\int_{x_{0}}p_{X_{0}\mymid X_{t}}(x_{0}\mymid x_{t})(x_{t}-\sqrt{\overline{\alpha}_{t}}x_{0})\mathrm{d}x_{0}\bigg\|_{2}\\
	& \qquad+\Big(\frac{1}{\sqrt{\alpha_t}(1-\overline{\alpha}_{t-1})}-\frac{1}{1-\overline{\alpha}_{t}}\Big)\int_{x_{0}}p_{X_{0}\mymid X_{t}}(x_{0}\mymid x_{t})\|x_{t}-\sqrt{\overline{\alpha}_{t}}x_{0}\|_2\mathrm{d}x_{0}\\
	& \qquad\qquad +\frac{1}{1-\overline{\alpha}_{t-1}}\int_{x_0} p_{X_{0}\mymid X_{t}}(x_{0}\mymid x_{t})\bigg\|g_t(x_t,\overline{\alpha}) - \sqrt{\frac{\overline{\alpha}}{\overline{\alpha}_t}}x_t\bigg\|_2\mathrm{d}x_0.
\end{align*}
Now applying the inequality~\eqref{eq:diff-s}, Lemma~\ref{lemma:x0}, and \eqref{eq:diff-x} on each term above separately, we achieve
\begin{align}
	&\Big\|\frac{1}{1-\overline{\alpha}_t}\int_{x_0}(x_t - \sqrt{\overline{\alpha}_t}x_{0})p_{X_0\mymid X_t}(x_0 \mymid x_t)\mathrm{d}x_0 - \frac{1}{1-\overline{\alpha}}\int_{x_0}(g_t(x_t,\overline{\alpha}) - \sqrt{\overline{\alpha}}x_{0})p_{X_0\mymid X_{\overline{\alpha}}}(x_0 \mymid g_t(x_t,\overline{\alpha}))\mathrm{d}x_0\Big\|_2\notag\\
	&\lesssim \frac{1}{1-\overline{\alpha}_t}\cdot\frac{d^{3/2}(1-\alpha_t)\log^{3/2}T}{(1-\overline{\alpha}_t)^{1/2}} + \frac{(1-\alpha_t)}{\sqrt{\alpha_t}(1-\overline{\alpha}_t)^2} \mathbb{E}\left[\big\| \sqrt{\overline{\alpha}_{t}}X_{0} - x_t \big\|_{2}\,\big|\,X_{t}=x_t\right]\notag\\
	&\qquad\qquad\qquad\qquad\qquad\qquad\qquad\qquad\qquad\qquad\qquad\qquad+ \frac{1}{1-\overline{\alpha}_t}\sup_{\overline{\alpha}_t<\overline{\alpha}<\overline{\alpha}_{t-1}}\Big\|\sqrt{\frac{\overline{\alpha}}{\overline{\alpha}_t}}x_t - g_t(x_t,\overline{\alpha})\Big\|_2\notag\\
	&\lesssim  \frac{d^{3/2}(1-\alpha_t)\log^{3/2}T}{(1-\overline{\alpha}_t)^{3/2}} + \frac{d^{1/2}(1-\alpha_t)\log^{1/2}T}{(1-\overline{\alpha}_t)^{3/2}} + \frac{d^{1/2}(1-\alpha_t)\log^{1/2}T}{(1-\overline{\alpha}_t)^{3/2}}\notag\\
	&\lesssim \frac{d^{3/2}(1-\alpha_t)\log^{3/2}T}{(1-\overline{\alpha}_t)^{3/2}}.\label{eq:diff-s4}
\end{align}
Finally, plugging inequalities~\eqref{eq:diff-s3} and \eqref{eq:diff-s4} into expression~\eqref{eq:A-diff} leads to 
\begin{align}
	\Big\|\frac{1}{(1-\overline{\alpha})^2}A_\alpha - \frac{1}{(1-\overline{\alpha}_t)^2}A_t\Big\|	
	\lesssim& \frac{d^{3/2}(1-\alpha_t)\log^{3/2}T}{(1-\overline{\alpha}_t)^{3/2}}\cdot \frac{d^{1/2}\log^{1/2}T}{(1-\overline{\alpha}_t)^{1/2}}\notag\\
	\lesssim& \frac{d^{2}(1-\alpha_t)\log^{2}T}{(1-\overline{\alpha}_t)^2}.\label{eq:diff-A}
\end{align} 
% \yuting{add explanations}\zh{done, the same as the analysis process for $A_t -A_{\overline{\alpha}}$, not repeat}

\paragraph{Controlling the second term.} 
With expression~\eqref{eq:diff-s2}, we can further control the quantity $\|\frac{1}{(1-\overline{\alpha})^2}B_\alpha - \frac{1}{(1-\overline{\alpha}_t)^2}B_t\|$.
By similar analysis, we can obtain 
\begin{align}
	&\Big\|\frac{1}{(1-\overline{\alpha})^2}B_\alpha - \frac{1}{(1-\overline{\alpha}_t)^2}B_t\Big\|\notag\\
	&\lesssim \frac{1}{\alpha_t (1-\overline{\alpha}_t)^2}\Big\|\int_{x_0}(x_t - \sqrt{\overline{\alpha}_t}x_{0})(x_t - \sqrt{\overline{\alpha}_t}x_{0})^{\top}p_{X_0\mymid X_t}(x_0 \mymid x_t)\mathrm{d}x_0\notag\\
	&\qquad\qquad\qquad\qquad- \int_{x_0}(x_t - \sqrt{\overline{\alpha}_t}x_{0})(x_t - \sqrt{\overline{\alpha}_t}x_{0})^{\top}p_{X_0\mymid X_{\overline{\alpha}}}(x_0 \mymid g_t(x,\overline{\alpha}))\mathrm{d}x_0\Big\|
	+ \frac{d(1-\alpha_t)\log T}{(1-\overline{\alpha}_t)^2}\notag\\
	&\lesssim \frac{d^{2}(1-\alpha_t)\log^{2}T}{(1-\overline{\alpha}_t)^2}.\label{eq:diff-B}
\end{align}

\paragraph{In summary,} taking the relations \eqref{eq:diff-A} and \eqref{eq:diff-B} collectively with inequality~\eqref{eqn:ochestra}, we obtain the following bound 
\begin{align*}
	\bigg\|\nabla s_{\overline{\alpha}}(g_t(x_t,\overline{\alpha})) - \nabla s_{t}(x_t)\bigg\| &\le \Big\|\frac{1}{(1-\overline{\alpha})^2}A_\alpha - \frac{1}{(1-\overline{\alpha}_t)^2}A_t\Big\| + \Big\|\frac{1}{(1-\overline{\alpha})^2}B_\alpha - \frac{1}{(1-\overline{\alpha}_t)^2}B_t\Big\|\notag\\ 
	&\lesssim \frac{d^{2}(1-\alpha_t)\log^{2}T}{(1-\overline{\alpha}_t)^2},
\end{align*}
which leads to the final result.

\subsection{Proof of Lemma~\ref{lemma:derivative-x}}
\label{sec:pf-lem-der-x}

The proof of this lemma is similar to that of Lemma~\ref{lemma:discrete}.
In particular, we shall prove this result by contradiction. 
Specifically, suppose that there exists $\overline{\alpha} \in [\overline{\alpha}_t, \overline{\alpha}_{t-1}]$ such that Lemma~\ref{lemma:derivative-x} does not hold.
Then, one can define
\begin{align*}
 	\widehat{\alpha} := \min\Big\{\overline{\alpha}\in [\overline{\alpha}_t, \overline{\alpha}_{t-1}] : \Big\|\frac{\partial g_t(x_t,\overline{\alpha})}{\partial x}-I \Big\| \gtrsim \frac{d(1-\alpha_t)\log T}{1-\overline{\alpha}_t} \Big\}.
 \end{align*}
With this definition of $\widehat{\alpha}$, it holds that for all $\overline{\alpha}_{t-1} \geq \overline{\alpha} > \widehat{\alpha}$, one has 
\begin{align}
\label{eqn:temp}
	\Big\|\frac{\partial g_t(x_t,\overline{\alpha})}{\partial x}\Big\| = 1 + O(d(1-\alpha_t)\log T).
\end{align}

Now consider the partial derivative of $g_t(x,\widehat{\alpha})$ at $\widehat{\alpha}$ where 
\begin{align*}
	\frac{\partial g_t(x,\widehat{\alpha})}{\partial x} - I_d &= \Big(\frac{1}{\sqrt{\alpha_{t}}}-1\Big)I_d + \frac{1}{2\sqrt{\alpha_{t}}}\int_{\overline{\alpha}_t}^{\widehat{\alpha}}\sqrt{\frac{\overline{\alpha}_t}{\overline{\alpha}^3}}\nabla s_{\overline{\alpha}}(g_t(x_t,\overline{\alpha}))\frac{\partial g_t(x_t,\overline{\alpha})}{\partial x}\mathrm{d}\overline{\alpha}. 
\end{align*}
The proof in Lemma~\ref{lemma:discrete} ensures that 
\begin{align*}
	\Big\|g_t(x_t,\overline{\alpha}) - \sqrt{\frac{\overline{\alpha}}{\overline{\alpha}_t}}x_t\Big\| \le c_5\sqrt{d(1-\alpha_t)\log T}
\end{align*}
for $(x_t,x_{t-1}) \in \mathcal{E}_t$. 
Thus, the analysis in the proof of (161a) in~\citet[Appendix C.1]{li2023towards} guarantees that
\begin{align}
	\big\|(1-\overline{\alpha})\nabla s_{\overline{\alpha}}(g_t(x_t,\overline{\alpha}))-I_d\big\| \lesssim d\log T,
\end{align}
which directly implies that
\begin{align*}
	\big\|\nabla s_{\overline{\alpha}}(g_t(x_t,\overline{\alpha}))\big\| \lesssim \frac{d\log T}{1-\overline{\alpha}_t},
\end{align*}

Combining these results together, we obtain
\begin{align*}
	\Big\|\frac{\partial g_t(x,\widehat{\alpha})}{\partial x} - I\Big\| 
	&\le \Big|\frac{1}{\sqrt{\alpha}_t} - 1\Big| + \frac{1}{2\sqrt{\alpha_{t}}}\int_{\overline{\alpha}_t}^{\widehat{\alpha}}\sqrt{\frac{\overline{\alpha}_t}{\overline{\alpha}^3}}\big\|\nabla s_{\overline{\alpha}}(g_t(x_t,\overline{\alpha}))\big\|\Big\|\frac{\partial g_t(x_t,\overline{\alpha})}{\partial x}\Big\|\mathrm{d}\overline{\alpha}\\
	&\le \Big|\frac{1}{\sqrt{\alpha}_t} - 1\Big| + \Big\|\frac{d\log T}{2\sqrt{\alpha_{t}}(1-\overline{\alpha}_t)}\int_{\overline{\alpha}_t}^{\overline{\alpha}_{t-1}}\sqrt{\frac{\overline{\alpha}_t}{\overline{\alpha}^3}}\mathrm{d}\overline{\alpha}\Big\|\\
	&\lesssim \frac{d(1-\alpha_t)\log T}{1-\overline{\alpha}_t},
\end{align*}
which contradicts the definition of $\widehat{\alpha}$.

\subsection{Proof of Lemma~\ref{lemma:J-Lip}}
\label{sec:pf-lem-J}

Define $\mathbb{S} \defn \{u \in \mathbb{R}^d : \|u\|_2 = 1\}$.
We first prove that for any $u \in \mathbb{S}^{d-1}$ and any $(x_t,x_{t-1}) \in \mathcal{E}_t$,
\begin{align}\label{eq:J-derivative}
	\big\|\nabla_x u^{\top}J_t(x_t)u\big\|_2 \lesssim \frac{1}{\sqrt{1-\overline{\alpha}_t}}d^{3/2}\log^{3/2} T,
\end{align}
where $\nabla_x u^{\top}J_t(x_t)u := \nabla_x u^{\top}J_t(x)u ~\big |_{x = x_t}.$
% \yuting{what is the definition of $\nabla_x u^{\top}J_t(x_t)u$; is it a number?}
% \zh{$\nabla_x u^{\top}J_t(x_t)u$ is the derivative of a scalar function with respect to $x_t$, thus a vector}

\paragraph{Proof of relation~\eqref{eq:J-derivative}.}
Recall that in Section~\ref{sec:jt-st}, we have shown that 
\begin{align*}
	s_t(x) &= -\frac{1}{1-\overline{\alpha}_t} \int_{x_0}(x - \sqrt{\overline{\alpha}_t}x_{0})p_{X_0\mymid X_t}(x_0 \mymid x)\mathrm{d}x_0,\\
	J_t(x) &= -(1-\overline{\alpha}_t)\frac{\partial s_t(x)}{\partial x}. 
\end{align*}
In view of these two relations and the definition of $J_{t}$, we can write $u^{\top}J_t(x_t)u$ as 
\begin{align*}
	u^{\top}J_t(x_t)u &= 1 +\frac{1}{1-\overline{\alpha}_{t}}\bigg\{\Big(\mathbb{E}\big[(X_{t}-\sqrt{\overline{\alpha}_{t}}X_{0})^{\top}u\mid X_{t}=x_t\big]\Big)^2
	-\mathbb{E}\Big[\big[\big(X_{t}-\sqrt{\overline{\alpha}_{t}}X_{0}\big)^{\top}u\big]_2^2\mid X_{t}=x_t\Big]\bigg\}.
\end{align*}
To further control $\nabla_x u^{\top}J_t(x_t)u$, let us consider the two terms on the right hand side separately. 

\begin{itemize}
	\item 
For the first term, one has 
\begin{align}
	\Big\|\nabla_{x_t} \Big(\mathbb{E}\big[(X_{t}-\sqrt{\overline{\alpha}_{t}}X_{0})^{\top}u\mid X_{t}=x_t\big]\Big)^2\Big\|_2 = & \Big\|\nabla_{x_t} \Big(\int_{x_0}\big[(x_t - \sqrt{\overline{\alpha}_t}x_{0})^{\top}u\big]p_{X_0\mymid X_t}(x_0 \mymid x_t)\mathrm{d}x_0\Big)^2\Big\|_2\notag\\
	=& \Big\|2(1-\overline{\alpha}_t)(s_t(x_t)^{\top}u)\cdot (1-\overline{\alpha}_t)\frac{\partial s_t(x_t)}{\partial x}\Big\|_2\notag\\
	\lesssim&(1-\overline{\alpha}_t)\|s_t(x_t)\|_2\cdot \Big\|(1-\overline{\alpha}_t)\frac{\partial s_t(x_t)}{\partial x}\Big\|.\label{eq:J-first-term}
\end{align}
By equation~\eqref{eqn:ice-cream}, we can compute that 
\begin{align}
	&\Big\|(1-\overline{\alpha}_t)\frac{\partial s_t(x_t)}{\partial x}\Big\|\notag\\
	&\le 1 + \frac{1}{1-\overline{\alpha}_t}\Big\|\int_{x_0}(x_t - \sqrt{\overline{\alpha}_t}x_{0})p_{X_0\mymid X_t}(x_0 \mymid x_t)\mathrm{d}x_0\Big(\int_{x_0}(x_t - \sqrt{\overline{\alpha}_t}x_{0})p_{X_0\mymid X_t}(x_0 \mymid x_t)\mathrm{d}x_0\Big)^{\top}\notag\\ 
	&\qquad \qquad - \int_{x_0}(x_t - \sqrt{\overline{\alpha}_t}x_{0})(x_t - \sqrt{\overline{\alpha}_t}x_{0})^{\top}p_{X_0\mymid X_t}(x_0 \mymid x_t)\mathrm{d}x_0\Big\|\notag\\
	&\lesssim 1 + \frac{1}{1-\overline{\alpha}_t}\Big\|\int_{x_0}(x_t - \sqrt{\overline{\alpha}}x_{0})p_{X_0\mymid X_t}(x_0 \mymid x_t)\mathrm{d}x_0\Big\|_2^2\notag\\
	&\le 1+ \frac{1}{1-\overline{\alpha}_t}\mathbb{E}_{X_0}\left[\big\| \sqrt{\overline{\alpha}_{t}}X_{0} - x_t \big\|^2_{2}\,\big|\,X_{t}=x_t\right]
	\lesssim d\log T.\label{eq:deriv-s-E}
\end{align}
Here, in the second inequality, we use the fact that for a column vector $Z \in \mathbb{R}^d$, we have
\begin{align*}
	\left\Vert \mathbb{E}\big[ZZ^{\top}\big]-\mathbb{E}[Z]\mathbb{E}[Z]^{\top}\right\Vert  & =\left\Vert \mathbb{E}\Big[\big(Z-\mathbb{E}[Z]\big)\big(Z-\mathbb{E}[Z]\big)^{\top}\Big]\right\Vert \leq\left\Vert \mathbb{E}\big[ZZ^{\top}\big]\right\Vert 
  \leq\mathbb{E}\big[\left\Vert ZZ^{\top}\right\Vert \big]=\mathbb{E}\big[\|Z\|_{2}^{2}\big],
\end{align*}
and the last line invokes Lemma~\ref{lemma:x0}.
Now plugging the bounds in inequality~\eqref{eq:s-E} and \eqref{eq:deriv-s-E} into inequality~\eqref{eq:J-first-term}, we obtain
\begin{align}
	\Big\|\nabla_{x_t} \Big(\mathbb{E}\big[(X_{t}-\sqrt{\overline{\alpha}_{t}}X_{0})^{\top}u\mid X_{t}=x_t\big]\Big)^2\Big\|_2
	\lesssim& (1-\overline{\alpha}_t)^{\frac{1}{2}}d^{3/2}\log^{3/2} T.\label{eq:J-derivative-1}
\end{align}

\item When it comes to the second term, some direct calculations give  
\begin{align}
	&\Big\|\nabla_{x_t} \mathbb{E}\Big[\big[\big(X_{t}-\sqrt{\overline{\alpha}_{t}}X_{0}\big)^{\top}u\big]_2^2\mid X_{t}=x_t\Big]\Big\|_2\notag\\
	=&\Big\|\nabla_{x_t} \int_{x_0}\big[(x_t - \sqrt{\overline{\alpha}_t}x_{0})^{\top}u\big]^2 p_{X_0\mymid X_t}(x_0 \mymid x_t)\mathrm{d}x_0\Big\|_2\notag\\
	\le&\Big\|2\int_{x_0}\big[(x_t - \sqrt{\overline{\alpha}_t}x_{0})^{\top}u\big]u\cdot p_{X_0\mymid X_t}(x_0 \mymid x_t)\mathrm{d}x_0\Big\|_2
	+ \Big\|\int_{x_0}\big[(x_t - \sqrt{\overline{\alpha}_t}x_{0})^{\top}u\big]^2 \frac{\partial}{\partial x_t}p_{X_0\mymid X_t}(x_0 \mymid x_t)\mathrm{d}x_0\Big\|_2\notag\\
	\le& 2\int_{x_0}\|(x_t - \sqrt{\overline{\alpha}_t}x_{0})\|_2\cdot p_{X_0\mymid X_t}(x_0 \mymid x_t)\mathrm{d}x_0
	+ \Big\|\int_{x_0}\big[(x_t - \sqrt{\overline{\alpha}_t}x_{0})^{\top}u\big]^2 \frac{\partial}{\partial x_t}p_{X_0\mymid X_t}(x_0 \mymid x_t)\mathrm{d}x_0\Big\|_2\notag\\
	\lesssim& (1-\overline{\alpha}_t)^{\frac{1}{3}}d^{1/2}\log^{1/2} T
	+ \Big\|\int_{x_0}\big[(x_t - \sqrt{\overline{\alpha}_t}x_{0})^{\top}u\big]^2 \frac{\partial}{\partial x_t}p_{X_0\mymid X_t}(x_0 \mymid x_t)\mathrm{d}x_0\Big\|_2,\label{eq:J-second-term-1}
\end{align}
where we use Lemma~\ref{lemma:x0} to obtain the last inequality.
To further bound the second term in inequality \eqref{eq:J-second-term-1}, we repeat the calculations for equations~\eqref{eq:J_ij} and \eqref{eq:partial-for-conditional} and deduce that
\begin{align}
	&\Big\|\int_{x_0}\big[(x_t - \sqrt{\overline{\alpha}_t}x_{0})^{\top}u\big]^2 \frac{\partial}{\partial x_t}p_{X_0\mymid X_t}(x_0 \mymid x_t)\mathrm{d}x_0\Big\|_2\notag\\
	\le& \Big\|\frac{1}{1-\overline{\alpha}_t}\int_{x_0}\big[(x_t - \sqrt{\overline{\alpha}_t}x_{0})^{\top}u\big]^2 (x_t - \sqrt{\overline{\alpha}_t}x_{0}) p_{X_0\mymid X_t}(x_0 \mymid x_t)\mathrm{d}x_0\Big\|_2\notag\\
	&\qquad + \Big\|\frac{1}{1-\overline{\alpha}_t}\int_{x_0}\big[(x_t - \sqrt{\overline{\alpha}_t}x_{0})^{\top}u\big]^2 p_{X_0\mymid X_t}(x_0 \mymid x_t)\mathrm{d}x_0\cdot\int_{x_0}(x_t - \sqrt{\overline{\alpha}_t}x_{0}) p_{X_0\mymid X_t}(x_0 \mymid x_t)\mathrm{d}x_0\Big\|_2\notag\\
	\lesssim& 
	\Big|\frac{1}{1-\overline{\alpha}_t}\mathbb{E}\big[\|x_t-\sqrt{\overline{\alpha}_t}x_{0}\|_2^3\mymid X_t = x_t\big]\Big| + \Big\|s_t(x_t)\mathbb{E}\big[\|x_t-\sqrt{\overline{\alpha}_t}x_{0}\|_2^2\mymid X_t = x_t\big]\Big\|_2 .\label{eq:J-second-term-2}
\end{align}
\end{itemize}
Taking colelctively the inequalities~\eqref{eq:J-second-term-1} and \eqref{eq:J-second-term-2}, we arrive at
\begin{align}
	&\nabla_{x_t} \mathbb{E}\Big[\big[\big(X_{t}-\sqrt{\overline{\alpha}_{t}}X_{0}\big)^{\top}u\big]_2^2\mid X_{t}=x_t\Big]\notag\\
	&\lesssim (1-\overline{\alpha}_t)^{\frac{1}{3}}d^{1/2}\log^{1/2} T + \Big|\frac{1}{1-\overline{\alpha}_t}\mathbb{E}\big[\|x_t-\sqrt{\overline{\alpha}_t}x_{0}\|_2^3\mymid X_t = x_t\big]\Big| + \Big\|s_t(x_t)\mathbb{E}\big[\|x_t-\sqrt{\overline{\alpha}_t}x_{0}\|_2^2\mymid X_t = x_t\big]\Big\|_2 \notag\\
	&\lesssim (1-\overline{\alpha}_t)^{\frac{1}{2}}d^{3/2}\log^{3/2} T.\label{eq:J-derivative-2}
\end{align}
where last inequality is a direct consequence of Lemma~\ref{lemma:x0}. 
Therefore, combining the two relations~\eqref{eq:J-derivative-1} and \eqref{eq:J-derivative-2}
yields the claimed relation~\eqref{eq:J-derivative}. 

\medskip

Next, we shall proceed to show that similar to the relation~\eqref{eq:J-derivative}, one also has 
\begin{align}
	\big\|\nabla_x u^{\top}J_t(x_{t-1}(\gamma))u\big\|_2 \lesssim \frac{1}{\sqrt{1-\overline{\alpha}_t}}d^{3/2}\log^{3/2} T, \label{eqn:shostakovich}
\end{align}
which holds for every $0 \le \gamma \le 1$.

\paragraph{Proof of inequality~\eqref{eqn:shostakovich}.} We make the observation that the derivations above to prove relation~\eqref{eq:J-derivative} only involves $X_t = x_t$ which satisfies the first condition in the definition of $\mathcal{E}_t$, namely, $-\log p_{X_t}(x_t) \leq c_3 d\log T$. 
% \yuting{where did you use; add details}\zh{done}
Now, let us prove that $-\log p_{X_t}(x_{t-1}(\gamma)) \le 2c_3 d\log T$ for $x_{t-1}(\gamma).$
Similar as in deriving inequality~\eqref{eq:second-term-1}, we can deduce 
% \yuting{why do we consider $\|x_{t-1} - \phi_t(x_t)\|_2$?}\zh{done}
\begin{align}
	\|x_{t-1} - \phi_t(x_t)\|_2 &\lesssim \Big\|x_{t-1} - \frac{x_t}{\sqrt{\alpha_t}}\Big\|_2 + \Big\|\Big(\frac{1}{\sqrt{\alpha}_t} - 1\Big)s_t(x_t)\Big\|_2 + \Big\|\frac{1}{2\sqrt{\alpha}_t}\int_{\overline{\alpha}_t}^{\overline{\alpha}_{t-1}}\sqrt{\frac{\overline{\alpha}_t}{\overline{\alpha}^3}}\big(s_{\overline{\alpha}}(g_t(x_t,\overline{\alpha})) - s_t(x_t)\big)\mathrm{d}\overline{\alpha}\Big\|_2\notag\\
	&\lesssim \sqrt{d(1-\alpha_t)\log T} + \sqrt{d(1-\alpha_t)^2\log T}\notag\\ 
	&\qquad\qquad\qquad\qquad+ \frac{1}{2\sqrt{\alpha_{t}}}\int_{\overline{\alpha}_t}^{\overline{\alpha}_{t-1}}\sqrt{\frac{\overline{\alpha}_t}{\overline{\alpha}^3}}\mathrm{d}\overline{\alpha}\Big(\sup_{\overline{\alpha}_t < \overline{\alpha} < \overline{\alpha}_{t-1}}\|s_{\widetilde{\alpha}}(g_t(x_t,\widetilde{\alpha})) - s_t(x_{t})\|_2\Big)\notag\\
	&\lesssim \sqrt{d(1-\alpha_t)\log T} + \sqrt{d(1-\alpha_t)^2\log T} + (1-\alpha_t)^2\bigg(\frac{d\log T}{1 - \overline{\alpha}_t}\bigg)^{3/2}\notag\\
	&\lesssim \sqrt{d(1-\alpha_t)\log T},\label{eq:xt-1-phi-xt-E}
\end{align}
where we use inequality~\eqref{eq:s-diff-2norm} in the third line. 
Since $x_{t-1}(\gamma) := \gamma x_{t-1} + (1-\gamma)\phi_{t}(x_t)$ and inequality~\eqref{eq:xt-1-phi-xt-E}, we can directly recognize that
\begin{align}
	\|x_{t-1}(\gamma) - \phi_t(x_t)\|_2\lesssim \sqrt{d(1-\alpha_t)\log T}.
\end{align}
Putting these two relations above together, it is easily seen that 
\begin{align}
	\label{eq:xt-1-gamma-phi-xt}
	\big\|x_{t-1} - x_{t-1}(\gamma)/\sqrt{\alpha_t} \big\| \leq c_4\sqrt{d(1-\alpha_t)\log T}.
\end{align}
% \yuting{check here}\zh{done}
In addition, in view of Lemma~\ref{lemma:river}, we know that for $(x_t, x_{t-1}) \in \mathcal{E}_t$ 
% \yuting{$\mathcal{E}_t$?}\zh{done} 
and any $\gamma \in [0,1]$, it holds that 
\begin{align}
	-\log p_{X_{t-1}}(x_{t-1}(\gamma)) \le 2c_3 d\log T \enspace\text{and}\enspace p_{X_{t-1}}(x) = \bigg(1+O\Big(\sqrt{\frac{d(1-\alpha_t)\log T}{1-\overline{\alpha}_t}}\Big)\bigg)p_{X_t}(x). \label{eq:log-density}
\end{align}
% Thus, for some pre-selected, large enough $c_3$, we still have
% \begin{align}
% 	-\log p_{X_t}(x_{t-1}(\gamma)) \le 2c_3 d\log T.
% \end{align}

From properties~\eqref{eq:xt-1-gamma-phi-xt} and \eqref{eq:log-density}, we conclude $(x_{t-1}(\gamma),x_{t-1}) \in \mathcal{E}_{t}$. It thus enables us to apply the same analysis as above on $J_t(x_{t-1}(\gamma))$, and draw the conclusion that
\begin{align}
	\big\|\nabla_x u^{\top}J_t(x_{t-1}(\gamma))u\big\|_2 \lesssim \frac{1}{\sqrt{1-\overline{\alpha}_t}}d^{3/2}\log^{3/2} T\notag
\end{align}
for $0 \le \gamma \le 1$. We complete the proof of the inequality~\eqref{eqn:shostakovich}.

\paragraph{In Summary.} 
Based on expression~\eqref{eq:J-derivative}, some direct calculations yield 
\begin{align} 
\label{eq:J-Lip}
	\|J_t(x_{t-1}) - J_t(\phi(x_t))\| \le \sup_{u \in \mathbb{S}^{d-1}} \Big|u^{\top}\Big(J_t(x_{t-1}) - J_t\big(\phi(x_t)\big)\Big)u\Big| \lesssim \frac{1}{\sqrt{1-\overline{\alpha}_t}}d^{3/2}\log^{3/2} T\|x_{t-1} - \phi(x_t)\|_2,
\end{align}
which concludes the proof of inequality~\eqref{eqn:J-lip-1}. 
In addition, as discussed after the inequality~\eqref{eq:phi-in-e}, the Lipschitz condition of $\Phi_{t-1 \to k}(x)$ allows us to prove $(x_k,x_{k-1}) \in \mathcal{E}_k$. 
Repeating the analysis above, we can conclude that
\begin{align*}
	\Big\|J_k\big(\Phi_{t-1 \to k}(x_{t-1}(\gamma))\big) - J_k\big(\Phi_{t \to k}(x_t)\big)\Big\| \lesssim \frac{1}{\sqrt{1-\overline{\alpha}_t}}d^{3/2}\log^{3/2} T\Big\|\Phi_{t-1 \to k}(x_{t-1}(\gamma)) - \Phi_{t \to k}(x_t)\Big\|_2,
\end{align*}
which thus completes the proof of inequality~\eqref{eqn:J-lip-2}.

%% file: Consistent_model_theory.bbl
\begin{thebibliography}{}

\bibitem[Anderson, 1982]{anderson1982reverse}
Anderson, B.~D. (1982).
\newblock Reverse-time diffusion equation models.
\newblock {\em Stochastic Processes and their Applications}, 12(3):313--326.

\bibitem[Benton et~al., 2023a]{benton2023linear}
Benton, J., De~Bortoli, V., Doucet, A., and Deligiannidis, G. (2023a).
\newblock Linear convergence bounds for diffusion models via stochastic
  localization.
\newblock {\em arXiv preprint arXiv:2308.03686}.

\bibitem[Benton et~al., 2023b]{benton2023error}
Benton, J., Deligiannidis, G., and Doucet, A. (2023b).
\newblock Error bounds for flow matching methods.
\newblock {\em arXiv preprint arXiv:2305.16860}.

\bibitem[Block et~al., 2020]{block2020generative}
Block, A., Mroueh, Y., and Rakhlin, A. (2020).
\newblock Generative modeling with denoising auto-encoders and {L}angevin
  sampling.
\newblock {\em arXiv preprint arXiv:2002.00107}.

\bibitem[Chen et~al., 2022a]{chen2022improved}
Chen, H., Lee, H., and Lu, J. (2022a).
\newblock Improved analysis of score-based generative modeling: User-friendly
  bounds under minimal smoothness assumptions.
\newblock {\em arXiv preprint arXiv:2211.01916}.

\bibitem[Chen et~al., 2023a]{chen2023probability}
Chen, S., Chewi, S., Lee, H., Li, Y., Lu, J., and Salim, A. (2023a).
\newblock The probability flow ode is provably fast.
\newblock {\em arXiv preprint arXiv:2305.11798}.

\bibitem[Chen et~al., 2022b]{chen2022sampling}
Chen, S., Chewi, S., Li, J., Li, Y., Salim, A., and Zhang, A.~R. (2022b).
\newblock Sampling is as easy as learning the score: theory for diffusion
  models with minimal data assumptions.
\newblock {\em arXiv preprint arXiv:2209.11215}.

\bibitem[Chen et~al., 2023b]{chen2023restoration}
Chen, S., Daras, G., and Dimakis, A.~G. (2023b).
\newblock Restoration-degradation beyond linear diffusions: A non-asymptotic
  analysis for {DDIM}-type samplers.
\newblock {\em arXiv preprint arXiv:2303.03384}.

\bibitem[De~Bortoli, 2022]{de2022convergence}
De~Bortoli, V. (2022).
\newblock Convergence of denoising diffusion models under the manifold
  hypothesis.
\newblock {\em arXiv preprint arXiv:2208.05314}.

\bibitem[De~Bortoli et~al., 2021]{de2021diffusion}
De~Bortoli, V., Thornton, J., Heng, J., and Doucet, A. (2021).
\newblock Diffusion {S}chr{\"o}dinger bridge with applications to score-based
  generative modeling.
\newblock {\em Advances in Neural Information Processing Systems},
  34:17695--17709.

\bibitem[Dhariwal and Nichol, 2021]{dhariwal2021diffusion}
Dhariwal, P. and Nichol, A. (2021).
\newblock Diffusion models beat gans on image synthesis.
\newblock {\em Advances in neural information processing systems},
  34:8780--8794.

\bibitem[Ding and Jin, 2023]{ding2023consistency}
Ding, Z. and Jin, C. (2023).
\newblock Consistency models as a rich and efficient policy class for
  reinforcement learning.
\newblock {\em arXiv preprint arXiv:2309.16984}.

\bibitem[Haussmann and Pardoux, 1986]{haussmann1986time}
Haussmann, U.~G. and Pardoux, E. (1986).
\newblock Time reversal of diffusions.
\newblock {\em The Annals of Probability}, pages 1188--1205.

\bibitem[Ho et~al., 2022]{ho2022imagen}
Ho, J., Chan, W., Saharia, C., Whang, J., Gao, R., Gritsenko, A., Kingma,
  D.~P., Poole, B., Norouzi, M., Fleet, D.~J., et~al. (2022).
\newblock Imagen video: High definition video generation with diffusion models.
\newblock {\em arXiv preprint arXiv:2210.02303}.

\bibitem[Ho et~al., 2020]{ho2020denoising}
Ho, J., Jain, A., and Abbeel, P. (2020).
\newblock Denoising diffusion probabilistic models.
\newblock {\em Advances in Neural Information Processing Systems},
  33:6840--6851.

\bibitem[Hyv{\"a}rinen, 2005]{hyvarinen2005estimation}
Hyv{\"a}rinen, A. (2005).
\newblock Estimation of non-normalized statistical models by score matching.
\newblock {\em Journal of Machine Learning Research}, 6(4).

\bibitem[Karras et~al., 2022]{karraselucidating}
Karras, T., Aittala, M., Aila, T., and Laine, S. (2022).
\newblock Elucidating the design space of diffusion-based generative models.
\newblock In {\em Advances in Neural Information Processing Systems},
  volume~35, pages 26565--26577.

\bibitem[Kim et~al., 2023]{kim2023consistency}
Kim, D., Lai, C.-H., Liao, W.-H., Murata, N., Takida, Y., Uesaka, T., He, Y.,
  Mitsufuji, Y., and Ermon, S. (2023).
\newblock Consistency trajectory models: Learning probability flow ode
  trajectory of diffusion.
\newblock {\em arXiv preprint arXiv:2310.02279}.

\bibitem[Kong et~al., 2020]{kong2020diffwave}
Kong, Z., Ping, W., Huang, J., Zhao, K., and Catanzaro, B. (2020).
\newblock Diffwave: A versatile diffusion model for audio synthesis.
\newblock {\em arXiv preprint arXiv:2009.09761}.

\bibitem[Lee et~al., 2023]{lee2023convergence}
Lee, H., Lu, J., and Tan, Y. (2023).
\newblock Convergence of score-based generative modeling for general data
  distributions.
\newblock In {\em International Conference on Algorithmic Learning Theory},
  pages 946--985.

\bibitem[Li et~al., 2024]{li2024accelerating}
Li, G., Huang, Y., Efimov, T., Wei, Y., Chi, Y., and Chen, Y. (2024).
\newblock Accelerating convergence of score-based diffusion models, provably.

\bibitem[Li et~al., 2023]{li2023towards}
Li, G., Wei, Y., Chen, Y., and Chi, Y. (2023).
\newblock Towards faster non-asymptotic convergence for diffusion-based
  generative models.
\newblock {\em arXiv preprint arXiv:2306.09251}.

\bibitem[Liu et~al., 2022]{liu2022let}
Liu, X., Wu, L., Ye, M., and Liu, Q. (2022).
\newblock Let us build bridges: Understanding and extending diffusion
  generative models.
\newblock {\em arXiv preprint arXiv:2208.14699}.

\bibitem[Lu et~al., 2022a]{lu2022dpm}
Lu, C., Zhou, Y., Bao, F., Chen, J., Li, C., and Zhu, J. (2022a).
\newblock {DPM-Solver}: A fast {ODE} solver for diffusion probabilistic model
  sampling in around 10 steps.
\newblock {\em Advances in Neural Information Processing Systems},
  35:5775--5787.

\bibitem[Lu et~al., 2022b]{lu2022dpmv2}
Lu, C., Zhou, Y., Bao, F., Chen, J., Li, C., and Zhu, J. (2022b).
\newblock {DPM-Solver++}: Fast solver for guided sampling of diffusion
  probabilistic models.
\newblock {\em arXiv preprint arXiv:2211.01095}.

\bibitem[Luhman and Luhman, 2021]{luhman2021knowledge}
Luhman, E. and Luhman, T. (2021).
\newblock Knowledge distillation in iterative generative models for improved
  sampling speed.
\newblock {\em arXiv preprint arXiv:2101.02388}.

\bibitem[Meng et~al., 2023]{meng2023distillation}
Meng, C., Rombach, R., Gao, R., Kingma, D., Ermon, S., Ho, J., and Salimans, T.
  (2023).
\newblock On distillation of guided diffusion models.
\newblock In {\em Proceedings of the IEEE/CVF Conference on Computer Vision and
  Pattern Recognition}, pages 14297--14306.

\bibitem[Nichol and Dhariwal, 2021]{nichol2021improved}
Nichol, A.~Q. and Dhariwal, P. (2021).
\newblock Improved denoising diffusion probabilistic models.
\newblock In {\em International Conference on Machine Learning}, pages
  8162--8171.

\bibitem[Pidstrigach, 2022]{pidstrigach2022score}
Pidstrigach, J. (2022).
\newblock Score-based generative models detect manifolds.
\newblock {\em arXiv preprint arXiv:2206.01018}.

\bibitem[Popov et~al., 2021]{popov2021grad}
Popov, V., Vovk, I., Gogoryan, V., Sadekova, T., and Kudinov, M. (2021).
\newblock Grad-tts: A diffusion probabilistic model for text-to-speech.
\newblock In {\em International Conference on Machine Learning}, pages
  8599--8608. PMLR.

\bibitem[Ramesh et~al., 2022]{ramesh2022hierarchical}
Ramesh, A., Dhariwal, P., Nichol, A., Chu, C., and Chen, M. (2022).
\newblock Hierarchical text-conditional image generation with {CLIP} latents.
\newblock {\em arXiv preprint arXiv:2204.06125}.

\bibitem[Rombach et~al., 2022]{rombach2022high}
Rombach, R., Blattmann, A., Lorenz, D., Esser, P., and Ommer, B. (2022).
\newblock High-resolution image synthesis with latent diffusion models.
\newblock In {\em IEEE/CVF Conference on Computer Vision and Pattern
  Recognition}, pages 10684--10695.

\bibitem[Salimans and Ho, 2022]{salimans2022progressive}
Salimans, T. and Ho, J. (2022).
\newblock Progressive distillation for fast sampling of diffusion models.
\newblock {\em arXiv preprint arXiv:2202.00512}.

\bibitem[Sohl-Dickstein et~al., 2015]{sohl2015deep}
Sohl-Dickstein, J., Weiss, E., Maheswaranathan, N., and Ganguli, S. (2015).
\newblock Deep unsupervised learning using nonequilibrium thermodynamics.
\newblock In {\em International Conference on Machine Learning}, pages
  2256--2265.

\bibitem[Song et~al., 2020]{song2020denoising}
Song, J., Meng, C., and Ermon, S. (2020).
\newblock Denoising diffusion implicit models.
\newblock {\em arXiv preprint arXiv:2010.02502}.

\bibitem[Song and Dhariwal, 2023]{song2023improved}
Song, Y. and Dhariwal, P. (2023).
\newblock Improved techniques for training consistency models.
\newblock {\em arXiv preprint arXiv:2310.14189}.

\bibitem[Song et~al., 2023]{song2023consistency}
Song, Y., Dhariwal, P., Chen, M., and Sutskever, I. (2023).
\newblock Consistency models.

\bibitem[Song and Ermon, 2019]{song2019generative}
Song, Y. and Ermon, S. (2019).
\newblock Generative modeling by estimating gradients of the data distribution.
\newblock {\em Advances in neural information processing systems}, 32.

\bibitem[Song and Ermon, 2020]{song2020improved}
Song, Y. and Ermon, S. (2020).
\newblock Improved techniques for training score-based generative models.
\newblock {\em Advances in neural information processing systems},
  33:12438--12448.

\bibitem[Song et~al., 2021]{song2020score}
Song, Y., Sohl-Dickstein, J., Kingma, D.~P., Kumar, A., Ermon, S., and Poole,
  B. (2021).
\newblock Score-based generative modeling through stochastic differential
  equations.
\newblock {\em International Conference on Learning Representations}.

\bibitem[Sun et~al., 2023]{sun2023accelerating}
Sun, W., Chen, D., Wang, C., Ye, D., Feng, Y., and Chen, C. (2023).
\newblock Accelerating diffusion sampling with classifier-based feature
  distillation.
\newblock In {\em 2023 IEEE International Conference on Multimedia and Expo
  (ICME)}, pages 810--815. IEEE.

\bibitem[Tang, 2023]{tang2023diffusion}
Tang, W. (2023).
\newblock Diffusion probabilistic models.
\newblock {\em preprint}.

\bibitem[Tang and Zhao, 2024]{tang2024contractive}
Tang, W. and Zhao, H. (2024).
\newblock Contractive diffusion probabilistic models.
\newblock {\em arXiv preprint arXiv:2401.13115}.

\bibitem[Vincent, 2011]{vincent2011connection}
Vincent, P. (2011).
\newblock A connection between score matching and denoising autoencoders.
\newblock {\em Neural computation}, 23(7):1661--1674.

\bibitem[Wang et~al., 2023]{wang2023videolcm}
Wang, X., Zhang, S., Zhang, H., Liu, Y., Zhang, Y., Gao, C., and Sang, N.
  (2023).
\newblock Videolcm: Video latent consistency model.
\newblock {\em arXiv preprint arXiv:2312.09109}.

\bibitem[Xue et~al., 2023]{xue2023sa}
Xue, S., Yi, M., Luo, W., Zhang, S., Sun, J., Li, Z., and Ma, Z.-M. (2023).
\newblock {SA-Solver}: Stochastic {Adams} solver for fast sampling of diffusion
  models.
\newblock {\em arXiv preprint arXiv:2309.05019}.

\bibitem[Zhang and Chen, 2022]{zhang2022fast}
Zhang, Q. and Chen, Y. (2022).
\newblock Fast sampling of diffusion models with exponential integrator.
\newblock {\em arXiv preprint arXiv:2204.13902}.

\bibitem[Zhao et~al., 2023]{zhao2023unipc}
Zhao, W., Bai, L., Rao, Y., Zhou, J., and Lu, J. (2023).
\newblock {UniPC}: A unified predictor-corrector framework for fast sampling of
  diffusion models.
\newblock {\em arXiv preprint arXiv:2302.04867}.

\end{thebibliography}
